\documentclass{vldb}
\usepackage{graphicx}
\usepackage{balance}  

\usepackage{braket}
\usepackage{amsmath}
\usepackage{amssymb}
\usepackage{bbm}

\usepackage{algorithm}
\usepackage{algpseudocode}

\usepackage{hyperref}
\usepackage{rotating}
\usepackage{subcaption}

\newtheorem{property}{Property}

\begin{document}


\title{Fast and Memory-Efficient Significant Pattern Mining via Permutation Testing}

\numberofauthors{4}

\author{
\alignauthor
Felipe Llinares L\'{o}pez\\
       \affaddr{D-BSSE, ETH Z{\"u}rich}\\
       \email{felipe.llinares@\\bsse.ethz.ch}
\alignauthor
Mahito Sugiyama\\
       \affaddr{ISIR, Osaka University}\\
       \affaddr{JST, PRESTO}\\
       \email{mahito@ar.sanken.\\osaka-u.ac.jp}
\alignauthor Laetitia Papaxanthos\\
       \affaddr{D-BSSE, ETH Z{\"u}rich}\\
       \email{laetitia.papaxanthos@\\bsse.ethz.ch}
\and  
\alignauthor Karsten M. Borgwardt\\
       \affaddr{D-BSSE, ETH Z{\"u}rich}\\
       \email{\hspace*{-20pt}karsten.borgwardt@bsse.ethz.ch}
}

\maketitle
 
\begin{abstract}
We present a novel algorithm, {\it Westfall-Young light}, for detecting patterns, such as itemsets and subgraphs, which are statistically significantly enriched in one of two classes. Our method corrects rigorously for multiple hypothesis testing and correlations between patterns
through the {\em Westfall-Young permutation procedure}, which empirically estimates the null distribution of pattern frequencies in each class via permutations.

In our experiments, {\it Westfall-Young light} dramatically outperforms the current state-of-the-art approach in terms of both runtime and memory efficiency on popular real-world benchmark datasets for pattern mining. The key to this efficiency is that unlike all existing methods, our algorithm neither needs to solve the underlying frequent itemset mining problem anew for each permutation nor needs to store the occurrence list of all frequent patterns. {\it Westfall-Young light} opens the door to significant pattern mining on large datasets that previously led to prohibitive runtime or memory costs.

\end{abstract}

\section{Introduction}
\label{sec:introduction}

Frequent pattern mining is one of the fundamental problems in data mining~\cite{Aggarwal14FPM}: In its most general form, one is given a database of transactions, each of which includes a set of items. A {\it frequent pattern} is then a set of items that co-occur in the same transactions more often than a predefined frequency threshold. Frequent pattern mining is at the heart of important problems such as association rule mining~\cite{Agrawal93}.

While the classic problem assumes that there is only one class of transactions, an important extension is to consider the case that two classes of transactions are given and one is interested in finding those sets of items that occur statistically significantly more often in one class than in the other. This problem of {\it significant pattern mining} is of fundamental importance to many applications of pattern mining, such as subgraph, substring and itemset mining:

 \textbf{Subgraph mining:} Given a collection of graphs, we aim to discover subgraphs which help separate the two classes of graphs. For instance, graphs represent drugs, which are either active or inactive regarding their ability to bind a specific target. Then, one would seek subgraphs, corresponding to molecular motifs, associated with drug activity~\cite{Takigawa13}. 

\textbf{Substring mining:} Similarly, objects could be labeled strings, leading to the problem of finding substrings which are characteristic of only one of the two string classes. For example, one might be looking for k-mers of DNA nucleotides which are enriched in a class of DNA strings known to contain binding sites for a protein of interest~\cite{Orenstein13}.

\textbf{Itemset mining:} Alternatively, in a labeled transaction database, one might be interested in identifying subsets of items whose occurrence is linked to the transaction class. One application of this setting is searching for higher-order interactions of binary predictors such as combinatorial regulation of gene expression by binding of transcription factors (TFs)~\cite{Lee12}. There, items correspond to binary features (TFs) and each transaction in the database is the subset of binary features taking value one for a specific realization of the binary predictors, labeled with the corresponding gene expression level (low or high).

Many methods for finding patterns that are {\it associated} with class membership have been proposed in the literature~\cite{Atzmueller06,Fan08,Novak09,Zimmermann10} (see~\cite{Zimmermann14} for a comprehensive survey), including several that provide a measure of statistical significance between the occurrence of a pattern and class membership~\cite{Arora14,Bay01,Hamalainen10,Yan08}. However, none of the aforementioned methods considers the inherent {\it multiple hypothesis testing problem} in significant pattern mining. It arises due to the combinatorial explosion in the number of patterns that one tests for significant association with class membership, which can lead to millions of patterns being deemed significant by mistake.
{\it Our goal in this article is to propose a method for significant pattern mining which can correct for multiple testing}.

The multiple hypothesis testing problem is that given a large enough number of tests, significant but false discoveries can be made with high probability~\cite{DudoitMT}. In pattern mining, where often billions of tests are being performed, it is very likely that millions of associations between patterns and class labels will appear to be statistically significant simply due to chance. 

Such false positive findings (patterns) are a problem for disciplines such as medical research, biology, neuroscience, psychology, or social sciences that use techniques of pattern mining for choosing patterns for further experimental investigation. Here, the cost of false positives is enormous, resulting in many hours of research and resources wasted. As a result, there is a pressing need to develop pattern mining algorithms which can rigorously correct for multiple testing and avoid false positives. 


 

In this paper, we introduce the first algorithm that is able to solve the multiple testing problem in significant pattern mining settings optimally and with attractive time and space requirements. By ``optimal' we refer to achieving the maximum possible number of true discoveries while strictly upper-bounding the probability of false discoveries by a predefined threshold. That is, we correct for multiple-testing by tightly controlling the Family Wise Error Rate (FWER, whose formal definition is given in the next section) without imposing any limit on the maximum pattern size.

The paper is organized as follows. Section~\ref{sec:background} describes the multiple hypothesis testing problem in detail, discusses the current state-of-the-art in significant pattern mining and introduces the necessary concepts for our proposal. Section~\ref{sec:proposal} introduces our contribution both from a theoretical and algorithmic perspective. Section~\ref{sec:experiments} presents an experimental study which empirically shows that our method outperforms existing techniques in two different setups of itemset mining and subgraph mining. For completeness, Section~\ref{sec:related_work} briefly describes other algorithms for significant pattern mining with sub-optimal FWER control. Finally, Section~\ref{sec:conclusions} summarizes our key findings.
\section{Background}
\label{sec:background}

\subsection{The Multiple Hypothesis Testing Problem}

In \emph{statistical association testing}, the dependence of each pattern on the class labels is quantified using its corresponding \emph{$p$-value}. In this context, the $p$-value is defined as the probability of measuring an association at least as extreme as the one observed in the data under the assumption that there is no real association present in the data-generating process. More precisely, let $H_{0}^{(i)}$ be the null hypothesis that the $i$-th pattern being examined is statistically independent of the class labels, $T^{(i)}$ the test-statistic chosen to evaluate association and $t^{(i)}$ the observed value of $T^{(i)}$. Assuming that larger values of $T^{(i)}$ indicate stronger dependence, the $p$-value is defined as $p_{i}=\mathrm{Pr}(\,T^{(i)} > t^{(i)} |  H_{0}^{(i)}\,)$. In practice, a pattern is considered to be significant if $p_{i} \le \alpha$, where $\alpha$ is called a \emph{significance threshold}. 

When only one pattern is being tested, the significance threshold $\alpha$ coincides with the \emph{type I error probability}; the probability of incorrectly rejecting a true null hypothesis. On the other hand, if we have a large number of patterns under study, statistical association testing becomes considerably more challenging because of the \emph{multiple hypothesis testing problem}. If there is a large number $D$ of patterns and those which satisfy $p_{i} \le \alpha$ are ruled as significant, then the probability that at least one out of the $D$ patterns will be wrongly deemed to be associated with the labels converges to $1$ at an exponential rate. That probability, commonly referred to as the \emph{Family-Wise Error Rate} (\emph{FWER}), can be written mathematically as $\mathrm{FWER}=\mathrm{Pr}\left(\mathrm{FP} > 0\right)$, where $\mathrm{FP}$ denotes the number of false positives incurred by the testing procedure, i.e., the number of patterns erroneously considered to be dependent on the labels. 

When designing a multiple hypothesis testing procedure, upper bounding the FWER by a predefined value $\alpha$ is the most popular approach. Note that, from a data mining perspective, $\alpha$ must never be considered as a parameter to be optimized but, rather, as an application-dependent requirement set before observing the data. Intuitively, $\alpha$ controls the amount of ``risk'' the user is willing to take when searching for patterns in the data. Larger values of $\alpha$ corresponding to more risk but, also, enhanced ability to discover real signals.

Then, the ideal problem to be solved is
\[
	\delta^{*} = \max\set{\delta | \mathrm{FWER}(\delta) \le \alpha},
\]
where $\delta$ is the \emph{corrected significance threshold} to determine which patterns are statistically significant and 
$\mathrm{FWER}(\delta)$ is the resulting FWER of that procedure. Nonetheless, since evaluating $\mathrm{FWER}(\delta)$ is a costly task, most methods use some quantity as a surrogate upper bound of the FWER instead. For instance, one of the most popular multiple testing corrections, the {\em Bonferroni correction}~\cite{Bonferroni36}, defines $f(\delta) = \delta D \ge \mathrm{FWER}(\delta)$. However, in data mining settings, it is intuitively clear that a large amount of patterns will in fact be a part of its super  patterns. Hence complex correlation structures involving the different $D$ test statistics will exist. Since the surrogate used by the Bonferroni correction completely neglects the dependence between test statistics, the resulting upper bound is quite loose, i.e., $\mathrm{FWER}(\delta) \ll \delta D$. That leads to over-conservative procedures with a significant loss of statistical power. It is therefore of great interest to develop algorithms to solve the original problem without using any upper bound on the FWER as a surrogate. 

\subsection{Westfall-Young Permutation-based\\ Hypothesis Testing}

By definition, the FWER can be expressed as $\mathrm{FWER}(\delta) = \mathrm{CDF}_{\Omega}(\delta)$, where $\Omega$ is the random variable $\Omega = \min_{i \in \mathcal{K}}\{\,p_{i} \mid H_{0}^{\mathcal{K}}\,\}$, $H_{0}^{\mathcal{K}} = \bigcap_{i \in \mathcal{K}}{H_{0}^{(i)}}$ and $\mathcal{K}$ is the set of true null hypotheses. Under a certain technical condition known as the subset pivotality condition, the FWER can also be obtained as the CDF of the random variable $\Omega^{\prime} = \min_{i \in \{\,1,\hdots,D\,\}}\{\,p_{i} \mid \mathbf{H}_{0}\,\}$, with $\mathbf{H}_{0} = \bigcap_{\set{1,\hdots,D}}{H_{0}^{(i)}}$, which is simpler as it does not depend on the (unknown) ground truth. Since most often there is no analytic formula available for the distribution of $\Omega^{\prime}$ nor its CDF, one resorts to resampling methods to approximate it.

In the context of statistical association testing, a permutation based resampling scheme proposed by Westfall and Young \cite{WYBook} is one of the most popular approaches to account for correlation structures. The idea is as follows: if one has to test the association of $D$ random variables $\left\{S_{i}\right\}_{i=1}^{D}$ with another random variable $Y$, a simple way of generating samples of the $p$-values under the global null (i.e.~to generate samples of $\Omega^{\prime}$) is by randomly shuffling (permuting) the observations of $Y$ with respect to those of the random variables $\left\{S_{i}\right\}_{i=1}^{D}$. The permuted datasets obtained that way are effective samples of the distribution $\mathrm{Pr}(\,\{S_{i}\}_{i=1}^{D}\,)\mathrm{Pr}(Y)$ and, therefore, correspond to samples obtained from the global null hypothesis that all variables $\left\{S_{i}\right\}_{i=1}^{D}$ are independent of $Y$. Those can then be used to obtain samples from $\mathrm{PDF}(\Omega^{\prime})$ by computing the respective $p$-values and taking the minimum across all patterns in the database, i.e.~$p_{\mathrm{min}}= \min_{i \in \set{1,\hdots,D}} \tilde{p}_{i} \sim \mathrm{PDF}(\Omega^{\prime})$, where $\tilde{p}_{i}$ is the $p$-value resulting from evaluating the association between $S_{i}$ and one realization of the permuted class labels $\widetilde{Y}$.

While conceptually simple, the Westfall-Young permutation scheme can be extremely computationally demanding. Generating a single sample from $\mathrm{PDF}(\Omega^{\prime})$ naively requires computing the $p$-values corresponding to all $D$ patterns using a specific permuted version of the class labels, $\widetilde{Y}$. Besides, to get a reasonably good empirical approximation to $\mathrm{FWER}(\delta) = \mathrm{CDF}_{\Omega^{\prime}}(\delta)$, one needs several samples $p_{\mathrm{min}}  \sim \mathrm{PDF}(\Omega^{\prime})$. As it will be shown in the experiments section, a number of samples in the order of $J=10^{3}$ or $J=10^{4}$ is needed to get reliable results. 

\subsection{State-of-the-art in Significant Pattern\\ Mining}

Terada et al.~\cite{TeradaIEEE} were the first to propose an efficient algorithm, called FastWY, which enables us to use Westfall-Young permutations in the setting of itemset mining.
The method uses inherent properties of discrete test statistics and succeeded to reduce the computational burden which the Westfall-Young permutation-based procedure entails. In the following we introduce the algorithm since we share the same problem setting and the key concept of the minimum attainable $p$-value.

In pattern mining, every $p$-value can be obtained from a $2 \times 2$ contingency table:
\begin{center}
	\begin{tabular}{|c|c|c|c|}
		\hline
		Variables & $S_{i}=1$ & $S_{i}=0$ & Row totals\\
		\hline
		$Y=1$ & $a_{i}$ & $n-a_{i}$ & $n$ \\
		\hline
		$Y=0$ & $x_{i}-a_{i}$ & $N+a_{i}-n-x_i$ & $N-n$ \\
		\hline
		Col totals & $x_{i}$ & $N-x_{i}$ & $N$ \\
		\hline
	\end{tabular}
\end{center}
Such tables are generally used to evaluate the strength of the association between two binary variables. In our case, those correspond to the class labels $Y$ and a variable $S_{i}$ that indicates whether the $i$-th pattern is present or not for each of the $N$ objects in the database ($S_i = 1$ meaning that the $i$-th pattern exists in the object). In a given database, $N$ denotes the total number of objects, $n$ the number of objects with class label $Y = 1$, $x_{i}$ the number of objects containing the $i$-th pattern, and $a_{i}$ the number of objects with class label $Y = 1$ containing the $i$-th pattern. Without loss of generality, we assume that the class labels are encoded so that $n \le N-n$.  

Let us use {\em Fisher's exact test}~\cite{FisherExactTest} to derive the test statistic, as it is one of the most popular methods to obtain a $p$-value out of a $2 \times 2$ contingency table. Fisher's exact test assumes that all marginals $x_{i}$, $n$, and $N$ are fixed. Thus, the table has only one degree of freedom left and one can model it as a one-dimensional count $a_{i}$. Under those assumptions, it can be shown that the underlying data-generating model under the null hypothesis of statistical independence between $S_{i}$ and $Y$ is a hypergeometric random variable. 

Mathematically, the probability is $\mathrm{Pr}(a_{i}=k | x_{i},n,N) = \binom{n}{k}\binom{N-n}{x_{i}-k} / \binom{N}{x_{i}}$ with $a_{i,\mathrm{min}} \leq k \leq a_{i,\mathrm{max}}$, where $a_{i,\mathrm{min}}=\max\{0,x_{i}+n-N\}$ and $a_{i,\mathrm{max}}=\min\{n,x_{i}\}$. If we observe $a_{i}=\gamma$, the corresponding \emph{one-tailed} $p$-value is obtained as $p_{i}(\gamma) = \min \{\Phi_{\mathrm{l}}(\gamma),\Phi_{\mathrm{r}}(\gamma)\}$, where we define $\Phi_{\mathrm{l}}(\gamma)=\sum_{k=a_{i,\mathrm{min}}}^{\gamma}{\mathrm{Pr}( a_{i}=k | x_{i},n,N )}$ and $\Phi_{\mathrm{r}}(\gamma)=\sum_{k=\gamma}^{a_{i,\mathrm{max}}}\mathrm{Pr}( a_{i}=k | x_{i},n,N )$ as the left and right tails of the hypergeometric distribution, respectively. To get a \emph{two-tailed} $p$-value, one can simply double the one-tailed $p$-value.

Since $a_{i}$ can only take $a_{i,\mathrm{max}} - a_{i,\mathrm{min}} +1$ different values, the corresponding $p$-values $p_{i}(a_{i})$ are also finitely many. Hence, there exists a {\em minimum attainable $p$-value} defined as $\Psi(x_{i},n,N) = \min\set{p_{i}(\gamma) | a_{i,\mathrm{min}} \leq \gamma \leq a_{i,\mathrm{max}}}$. As $n$ and $N$ depend only on the labels and are therefore fixed for all patterns, we abbreviate $\Psi(x_{i},n,N)$ as $\Psi(x_{i})$ in the rest of the paper. Related to the minimum attainable $p$-value, we also introduce the \emph{set of testable patterns at corrected significance level $\delta$},  $\mathcal{I}_{T}(\delta) = \{\,i \in \left\{1,\hdots,D\right\} \mid \Psi(x_{i}) \le \delta\,\}$. Thus patterns not in $\mathcal{I}_{T}(\delta)$ can never be significant at level $\delta$.

The key to the FastWY algorithm is to exploit this minimum attainable $p$-value, which the algorithm requires to be monotonically decreasing in $[0,N]$. To achieve this requirement, FastWY uses a surrogate lower bound $\hat{\Psi}(x_{i})$ such that
\begin{flalign*}
\hat{\Psi}(x_{i}) = \begin{cases}\Psi(x_{i}) & 0 \leq x_{i} \leq n, \\1 \big/ \binom{N}{n} & n < x_{i} \leq N \\\end{cases}
\end{flalign*}
and defines $\hat{\mathcal{I}}_{T}(\delta) = \{\,i \in \left\{1,\hdots,D\right\} \mid \hat{\Psi}(x_{i}) \le \delta\,\}$, which always satisfies $\mathcal{I}_{T}(\delta) \subset \hat{\mathcal{I}}_{T}(\delta)$. In other words, the set of patterns they retrieve might contain some unnecessary patterns.
Nonetheless, as a result of the monotonicity of $\hat{\Psi}$, we can rewrite $\hat{\mathcal{I}}_{T}(\delta)$ as $\hat{\mathcal{I}}_{T}(\sigma(\delta)) = \{\,i \in \left\{1,\hdots,D\right\} \mid x_{i} \ge \sigma(\delta)\,\}$ using $\sigma(\delta)=\inf \{\,x_i \mid \hat{\Psi}(x_{i}) \le \delta\,\}$. This new formulation is important because $\hat{\mathcal{I}}_{T}(\sigma(\delta))$ can be readily seen to correspond to an instance of \emph{frequent pattern mining with support $\sigma(\delta)$}, a well studied problem in data mining.

That property is the basis of FastWY as follows:
If the $p$-values for all patterns in $\hat{\mathcal{I}}_{T}(\sigma)$ are known and $p^{\prime}_{\mathrm{min}}=\min\set{\tilde{p}_{i} | i \in \hat{\mathcal{I}}_{T}(\sigma)} \le \hat{\Psi}(\sigma)$, no pattern in $\set{1,\hdots,D} \setminus \hat{\mathcal{I}}_{T}(\sigma)$ can possibly attain a $p$-value smaller than $p^{\prime}_{\mathrm{min}}$. Hence $p_{\mathrm{min}}=p^{\prime}_{\mathrm{min}}$ and the search can be stopped early. That can be exploited to sample from $\mathrm{PDF}(\Omega^{\prime})$ without having to explicitly compute all $D$ $p$-values. 

FastWY is based on a decremental search scheme starting with the support $\sigma=n$. For each $\sigma$, first a frequent pattern miner is used as a black box to retrieve the set $\hat{\mathcal{I}}_{T}(\sigma)$. Then, the $p$-values $\tilde{p}_{i}$ are computed for all $i \in \hat{\mathcal{I}}_{T}(\sigma)$ and $p^{\prime}_{\mathrm{min}}=\min\set{\tilde{p}_{i} | i \in \hat{\mathcal{I}}_{T}(\sigma)}$ is evaluated. If $p^{\prime}_{\mathrm{min}} \le \hat{\Psi}(\sigma)$, no other pattern makes $p^{\prime}_{\mathrm{min}}$ smaller and, therefore, $p^{\prime}_{\mathrm{min}}=p_{\mathrm{min}}$ constitutes an exact sample from $\mathrm{PDF}(\Omega^{\prime})$. Otherwise if $p^{\prime}_{\mathrm{min}} > \hat{\Psi}(\sigma)$, $\sigma$ is decreased by one unit and the whole procedure is repeated until the condition $p^{\prime}_{\mathrm{min}} \le \hat{\Psi}(\sigma)$ is satisfied.

If $J$ permutations are needed to empirically estimate the FWER, the procedure has be to repeated $J$ times. That includes the whole sequence of frequent mining problems needed to retrieve the sets $\hat{\mathcal{I}}_{T}(\sigma)$ for each support value $\sigma$ used throughout the decremental search. Given the usual range of values for $J$, such an approach is just as unfeasible in practice as the original brute force baseline.

Interestingly, careful inspection of the code kindly shared by the authors in their website reveals that the actual implementation of the algorithm is different from the description in \cite{TeradaIEEE}. Indeed, to alleviate the inadmissible burden of repeating the whole frequent pattern mining process $J$ times, the authors resort to storing the realizations of the variables $S_{i}$ for all $i \in \hat{\mathcal{I}}_{T}(\sigma)$ and every value of $\sigma$ explored during the decremental search. Clearly, this decision corresponds to a drastic trade-off between runtime and memory usage since, as we will show empirically, there exist many databases for which the amount of information to be stored when following that approach is too large to fit even in large servers equipped with 512~GB of RAM. Furthermore, we will also demonstrate that the algorithm resulting from trading off memory in exchange for runtime is still slow even for mid-sized datasets and can be greatly improved.

When referring to the FastWY algorithm in \cite{TeradaIEEE} throughout the remaining of this paper, we mean the runtime optimized version of the algorithm, given that the algorithm exactly as described in \cite{TeradaIEEE} simply cannot be run in a reasonable amount of time.
\section{The Westfall-Young Light\\ Approach}
\label{sec:proposal}

As discussed previously, despite being an interesting idea, several properties of the FastWY algorithm do not allow its application on large datasets. The main reasons why the state-of-the-art algorithm is inefficient are: (1) It uses a decremental search strategy, which is known to be orders of magnitude slower than incremental search~\cite{Minato} (see Section~\ref{sec:related_work} for more detail); (2) naive application of the method requires either repeating pattern mining $J$ times, with $J$ in the order of $10^{4}$, or alternatively storing in memory the list of occurrences among the $N$ objects in the database of all frequent patterns; (3) $J$ exact samples from $\mathrm{PDF}(\Omega^{\prime})$ have to be computed, even though the calculation of $\delta^{*}$ actually involves only the smallest $\lceil \alpha J \rceil$ samples; (4) as a direct consequence of (3), the larger $J$ the more likely it is that some of the samples of $\Omega^{\prime}$ fall in the upper tail of the distribution, hence requiring to mine patterns with unnecessarily low supports and significantly increasing both the overall runtime and the memory requirements; (5) it relies on using a surrogate lower bound $\hat{\Psi}(x_{i})$ on the minimum attainable $p$-value $\Psi(x_{i})$, despite the fact that such a strategy might consider some unnecessary patterns.

In this section, we show how to remove all the limitations listed above, resulting in a novel procedure for permutation-based significant pattern mining. This new algorithm requires significantly less memory and can be shown to be up to 3 orders of magnitude faster in real world data. 

We show first how to get rid of limitation (5). Then, after introducing two previously unexploited key properties of the FWER estimator used in permutation testing, we describe how to get rid of limitations (1) up to (4).

\subsection{Removing Limitation (5)}

\begin{figure}[t]
\centering
\includegraphics[width=.75\linewidth]{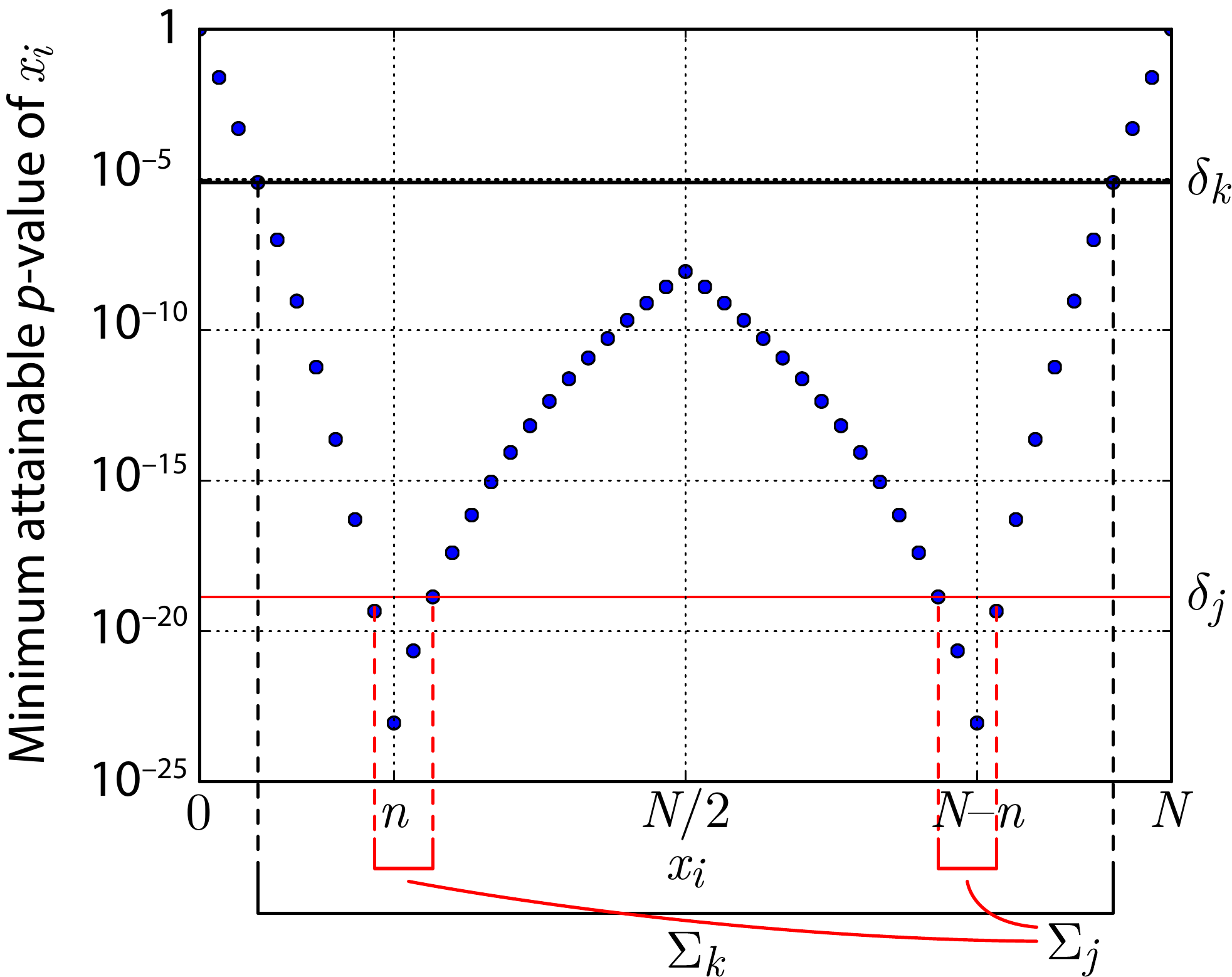}
\caption{Exact minimum attainable $p$-value $\Psi(x_{i})$ for $n=10$ and $N=50$ (blue dots).
Two different types of regions $\Sigma_j$ and $\Sigma_k$ are illustrated (red and black lines).}
\label{fig:Exact_psi}
\end{figure}

Discrete test statistics not only result in a finite set of $p$-values but, similarly, they have finitely many minimum attainable $p$-values. The exact minimum attainable $p$-value function $\Psi(x_{i})$ is given by: \begin{equation}
\label{eq:exactpsi}
	\Psi(x_{i}) = \begin{cases}
		\frac{n!}{N!}\frac{(N-x_{i})!}{(n-x_{i})!}   & 0 \leq x_{i} \leq n, \\
		\frac{(N-n)!}{N!}\frac{x_{i}!}{(x_{i}-n)!}    & n < x_{i} \leq \frac{N}{2}, \\	
		\frac{(N-n)!}{N!}\frac{(N-x_{i})!}{((N-n)-x_{i})!}    & \frac{N}{2} < x_{i} \leq N-n, \\
		\frac{n!}{N!}\frac{x_{i}!}{(x_{i}-(N-n))!}   & N-n \leq x_{i} \leq N, \\
	\end{cases}
\end{equation}
(see Figure \ref{fig:Exact_psi}).
Note that $\Psi(x_{i}) = \Psi(N - x_{i})$ always folds for $0 \le x_{i} \le N/2$. 
The function $\Psi(x_{i})$ takes $\lfloor \frac{N}{2} \rfloor + 1$ different values in $[0,N]$. This in turn implies that, despite depending on a real-valued parameter $\delta \in [0,1]$, there are only $\lfloor \frac{N}{2} \rfloor + 1$ different sets of testable patterns $\mathcal{I}_{T}(\delta)$. If we sort the range of $\Psi(x_{i})$ as a monotonically increasing sequence $\delta_{\lfloor \frac{N}{2} \rfloor} < \delta_{\lfloor \frac{N}{2} \rfloor -1} < \hdots < \delta_{1} < \delta_{0}=1$, it becomes clear that $\mathcal{I}_{T}(\delta)=\mathcal{I}_{T}(\delta_{k})$ for all $\delta \in [\delta_{k},\delta_{k-1})$. More importantly, for each threshold $\delta_{k}$ with $k=0,\hdots,\lfloor \frac{N}{2} \rfloor$, there exists a corresponding region $\Sigma_{k} \in [0,N]$ such that $x_{i} \in \Sigma_{k}$ if and only if $i \in \mathcal{I}_{T}(\delta_{k})$. Those regions can be of two different types: (1) if $\delta_{k} < \Psi(\lfloor \frac{N}{2} \rfloor)$, the region is the union of two symmetric intervals, i.e., $ \Sigma_{k} = [\sigma_{l}^{k},\sigma_{u}^{k}] \cup [N-\sigma_{u}^{k},N-\sigma_{l}^{k}]$; (2) if $\delta_{k} \ge \Psi(\lfloor \frac{N}{2} \rfloor)$ the region is composed of a single interval, $\Sigma_{k} = [\sigma_{l}^{k},N-\sigma_{l}^{k}]$. We have introduced $\sigma_{l}^{k}$ and $\sigma_{u}^{k}$ as $\sigma_{l}^{k} = \min\set{x \in [0,n] | \Psi(x) \le \delta_{k}}$ and $\sigma_{u}^{k} = \max\set{x \in [n,\lfloor\frac{N}{2}\rfloor] | \Psi(x) \le \delta_{k}}$, respectively. Note also that, regardless of the type, the regions $\Sigma_{k}$ are always symmetric around $N / 2$. Both types of regions are illustrated with an example in Figure \ref{fig:Exact_psi}.

Finally, the last key observation is that $\Sigma_{j} \subset \Sigma_{k}$ whenever $j > k$. In other words, the corresponding regions $\set{x | \Psi(x) \le \delta}$ shrink monotonically with respect to $\delta$. That, and not the strict monotonicity of $\Psi(x_{i})$ as argued in \cite{TeradaPNAS}, is the actual property needed to implement the method in \cite{Tarone} via branch-and-bound techniques. Indeed, recovering $\mathcal{I}_{T}(\delta_{k})$ amounts to an instance of frequent pattern mining with support $\sigma_{l}^{k}$ in which frequent patterns whose support is not in $\Sigma_{k}$ are discarded. Besides, given $\Sigma_{k}$ and $\delta_{k}$, computing $\Sigma_{k+1}$ and $\delta_{k+1}$ has complexity $O(1)$ as we will discuss later.

\subsection{Removing Limitations (1)-(4)}

Now we show how, by exploiting properties of the FWER estimator, it is possible to rearrange computations in a way that allows: (1) using an incremental search scheme, so that the frequent pattern mining algorithm needs to be run only once instead of $J$ times without any extra memory requirements other than just storing the binary $N$-by-$J$ matrix of permuted class labels (limitations 1 and 2); (2) only the minimum $\lceil \alpha J \rceil$ smallest samples from $\mathrm{PDF}(\Omega^{\prime})$ need to be generated exactly, significantly reducing the overall frequent pattern mining effort (limitations 3 and 4).

Let $p_{\mathrm{min}}^{(j)}, \; j=1,\hdots,J$ be the set of minimum $p$-values for each permutation, i.e. $J$ different exact samples from the distribution $\mathrm{PDF}(\Omega^{\prime})$. Then, the empirical estimator of the FWER at corrected significance level $\delta$ is given as
\[
\mathrm{FWER}(\delta)=\frac{1}{J}\sum_{j=1}^{J}{\mathbbm{1}\left[p_{\mathrm{min}}^{(j)} \le \delta\right]},
\]
where $\mathbbm{1}[\bullet]$ denotes a function which evaluates to $1$ if its argument is true and to $0$ otherwise.
The following two properties of that estimator are the theoretical basis of our contribution:

\begin{property}
\label{prop:FWER}
	Whenever a new pattern $i$ is processed, the updated empirical FWER estimate can never decrease. 
\end{property}
\begin{proof}
	We have $\min\set{\tilde{p}_{i} | i \in \mathcal{I} \cup \set{k}}=\min\{\min\{\tilde{p}_{i} \mid i \in \mathcal{I}\}, \tilde{p}_{k}\}$ and hence $\mathbbm{1} [\,\min\{\tilde{p}_{i} \mid i \in \mathcal{I}\} \le \delta_{k}\,] \le \mathbbm{1} [\,\min\{\tilde{p}_{i} \mid i \in \mathcal{I} \cup \set{k}\} \le \delta_{k}\,]$. Thus this property readily follows.
\end{proof}
\begin{property}
\label{prop:eval}
	$\mathrm{FWER}(\delta)$ for all $\delta \in [0,\delta_{k}]$ can be evaluated exactly using only the $p$-values of patterns in $\mathcal{I}_{T}(\Sigma_{k})$. 
\end{property}
\begin{proof}
	Let $p_{\mathrm{min}}^{\prime} = \min\{\,\tilde{p}_{i} \mid i \in \mathcal{I}_{T}(\Sigma_{k})\,\}$. By definition of $\mathcal{I}_{T}(\Sigma_{k})$, if $p_{\mathrm{min}}^{\prime} > \delta_{k}$ then $p_{\mathrm{min}} > \delta_{k}$, leading to $\mathbbm{1} [\,p_{\mathrm{min}} \le \delta_{k}\,] = \mathbbm{1} [\,p_{\mathrm{min}}^{\prime} \le \delta_{k}\,]$.
\end{proof}
By putting both properties together, in Section~\ref{subsec:ouralgorithm} we will show how one can implement an incremental counterpart of FastWY which removes all limitations (1) up to (4).

\subsection{Evaluating Fisher's Exact Test p-values in Negligible Time}

Here we provide an efficient technique for computing $p$-values of Fisher's exact test, which is also crucial for improving the efficiency if $J$ is large.

Roughly speaking, one can divide the computational effort required to apply the Westfall-Young permutation testing procedure to data mining problems into 4 main categories: (1) compute the matrix of permuted class labels; (2) for each pattern, compute the cell counts $a_{i}^{j}$ for all $J$ permutations; (3) for each pattern, compute the $p$-values corresponding to the $J$ cell counts, $\tilde{p}_{i}^{(j)}(a_{i}^{(j)})$ and; (4) the frequent pattern mining itself. (1) can be considered to be negligible in most large datasets and therefore is not a major concern. (4) will be reduced to the bare minimum by using an incremental search strategy. While that effort might still be considerable, it appears difficult a priori to reduce it further. Finally, both (2) and (3) have a complexity $O(x_{i}J)$ per pattern. Even if that might be small, bearing in mind that the number of patterns to be inspected can be in the order of trillions and $J \approx 10^{4}$, it could be an even more demanding task than frequent pattern mining itself, depending on the dataset. Nonetheless, by using a careful arrangement of the $p$-value computations only made possible by incremental search, we will show how to reduce the complexity of (3) from $O(x_{i}J)$ per pattern to $O(x_{i})$ per pattern, leaving (2) and (4) as the sole contributors to the overall runtime in our algorithm. For that we exploit the following computational property of Fisher's exact test:

\begin{property}
\label{prop:compute_p_value}
	For fixed $x_{i}$, $n$, and 	$N$, the computational complexity of evaluating Fisher's exact test $p$-value $p_{i}(\gamma)$ for a single value of $\gamma$ or for all possible values of $\gamma$ in $[a_{i,\mathrm{min}},a_{i,\mathrm{max}}]$ is the same and equal to $O(\min\{x_{i},n\})$.
\end{property}
\begin{proof}
	Let $p_{i}(\gamma) = \min \left(\Phi_{\mathrm{l}}(\gamma),\Phi_{\mathrm{r}}(\gamma)\right)$, with $\Phi_{\mathrm{l}}(\gamma)$ and $\Phi_{\mathrm{r}}(\gamma)$ as defined in section \ref{sec:background}. Computing $p_{i}(\gamma)$ for a single value of $\gamma$ requires evaluating the probability mass of a hypergeometric random variable with parameters $x_{i}$, $n$, $N$ in all its support and computing the left and right tail mass, with split point at $\gamma$. Hence, the complexity of evaluating $p_{i}(\gamma)$ for a single fixed $\gamma$ is $O(\min\{x_{i},n\})$. Note that all of those computations can be shared across all $J$ permutations as long as $x_{i}$, $n$, $N$ are fixed. We can therefore compute the $p$-values corresponding to all possible values of $\gamma$ instead of just a single one also with the same complexity $O(\min\{x_{i},n\})$. 
\end{proof}

To exploit this property, we use an enumeration scheme which processes each pattern at a time but all $J$ permutations simultaneously, something which is not feasible with the decremental scheme of FastWY. As a result, by using Property~\ref{prop:compute_p_value}, we share the computational burden of computing $p$-values for a given pattern across all $J$ permutations, in a way that the resulting complexity is the same as for the case $J=1$. In contrast, FastWY resorts to the strategy of storing in memory all previously evaluated $p$-values to avoid recomputing them many times. Again, this is an implementation aspect not discussed in their paper but present in their code, which represents another memory versus runtime trade-off.

\subsection{The Algorithm: Westfall-Young Light}
\label{subsec:ouralgorithm}

\begin{algorithm}[ht]
\begin{small}
\caption{Westfall-Young light}\label{alg:incfastwy}
 \begin{algorithmic}[1]
  \State {\bfseries Input:} Dataset, class labels $\mathbf{y}$, number of permutations $J$, and significance threshold $\alpha$
  \State {\bfseries Output:} Corrected significance threshold $\delta^*$
	\Function{\scshape Westfall-Young Light}{$\alpha$, $J$}
		\For{$j = 1,\hdots,J$}
			\State $\mathbf{y}^{(j)} \leftarrow \mathrm{permute}(\mathbf{y})$ // Permute class labels $\mathbf{y}$
			\State $p_{\mathrm{min}}^{(j)} \leftarrow 1$
		\EndFor
		\State $k \leftarrow 1$, $\delta_{k} \leftarrow n / N$
		\State $\sigma^{k}_{l} \leftarrow 1$, $\sigma^{k}_{u} \leftarrow \lfloor N / 2\rfloor$
                \State $\mathrm{flag} \leftarrow 1$
 		\State {\scshape ProcessNext}$(\mathrm{root}, N)$
		\State // Patterns are enumerated through a rooted tree
	\EndFunction
	\Function{\scshape ProcessNext}{$i$, $x_{i}$}
		\If{$x_{i} \in [\sigma_{l}^{k},\sigma_{u}^{k}] \cup [N-\sigma_{u}^{k},N-\sigma_{l}^{k}]$}
			\State Compute $p$-values $p_{i}(\gamma)$ for all $\gamma \in [a_{i,\mathrm{min}},a_{i,\mathrm{max}}]$
			\For{$j = 1,\hdots,J$}
				\State Compute $a_{i}^{(j)}$
				\State $p_{\mathrm{min}}^{(j)} \leftarrow \min\{\,p_{\mathrm{min}}^{(j)}, p_{i}(a_{i}^{(j)})\,\}$
			\EndFor
			\While{$\frac{1}{J}\sum_{j=1}^{J}{\mathbbm{1}\left[p_{\mathrm{min}}^{(j)} \le \delta_{k}\right]} > \alpha$}
				\State $k \leftarrow k +1$
				\State {\scshape UpdateThreshold}$(k,\delta_{k},\sigma^{k}_{l},\sigma^{k}_{u})$
			\EndWhile
		\EndIf
		\For{$j \in \mathrm{Children}(i)$}
			\State Compute $x_{j}$
			\If{$x_{j} \ge \sigma^{k}$}
				\State {\scshape ProcessNext}$(j,x_{j})$
			\EndIf
		\EndFor
		\State Return $\delta^{*} \in [\delta_{k},\delta_{k-1})$
	\EndFunction
	\Function{\scshape UpdateThreshold}{$k, \delta_{k}, \sigma^{k}_{l}, \sigma^{k}_{u}$}
		\If{$\mathrm{flag} = 1$}
			\State $\sigma^{k}_{l} \leftarrow \sigma^{k}_{l} + 1$
			\If{$\Psi(\sigma^{k}_{l}) \ge \Psi(\sigma^{k}_{u})$}
				\State $\delta_{k} \leftarrow \Psi(\sigma^{k}_{l})$
			\Else
				\State $\delta_{k} \leftarrow \Psi(\sigma^{k}_{u})$
				\State $\mathrm{flag} \leftarrow 0$
			\EndIf
		\Else
			\State $\sigma^{k}_{u} \leftarrow \sigma^{k}_{u} - 1$
			\If{$\Psi(\sigma^{k}_{l}) \ge \Psi(\sigma^{k}_{u})$}
				\State $\delta_{k} \leftarrow \Psi(\sigma^{k}_{l})$
				\State $\mathrm{flag} \leftarrow 1$
			\Else
				\State $\delta_{k} \leftarrow \Psi(\sigma^{k}_{u})$
			\EndIf
		\EndIf

	\EndFunction
 \end{algorithmic}
 \end{small}
 \end{algorithm}

Exploiting all those properties, we propose the following algorithm that removes all limitations (1) up to (4) equipped with efficient computation of $p$-values.

First, precompute all $J$ sets of permuted class labels. Then, initialize $\delta_k$, $\Sigma_k$, and $\sigma_k$ for $k=1$, that is, the threshold to the maximum (non-trivial) value $\delta_{1}=n / N$, the associated testable region $\Sigma_{1}=[1,N-1]$, and the minimum support for frequent pattern mining $\sigma_{l}^{1} = 1$. Note that we skipped $k=0$, since it corresponds to a trivial threshold $\delta_{0}=1$ for which every pattern is testable and significant. Next, start enumerating patterns using the frequent pattern miner of choice and, every time a frequent pattern is found, do the following: (1) check if its support $x_{i}$ satisfies $x_{i} \in \Sigma_{k}$ (i.e. check if the pattern is testable), if not simply continue mining; (2) if yes, precompute all possible $p$-values for a hypergeometric random variable with parameters $x_{i}$, $n$ and $N$; (3) compute the cell counts $a_{i}^{j}$ for all $j=1,\hdots,J$ and fetch the corresponding $p$-values $\tilde{p}_{i}^{j}=p_{i}(a_{i}^{(j)})$, updating $p_{\mathrm{min}}^{(j)}$ if $\tilde{p}_{i}^{j} < p_{\mathrm{min}}^{(j)}$; (4) check if the current FWER is too high, if it is, increase $k$ (thus reducing $\delta_{k}$, shrinking $\Sigma_{k}$, and increasing the minimum support $\sigma_{l}^{k}$ for frequent pattern mining) until the FWER is below the target $\alpha$ again; and (5) continue running the frequent pattern miner without restarting.

This combined scheme of frequent pattern enumeration and adaptive threshold adjustment continues until all patterns for a certain $(k,\delta_{k},\Sigma_{k})$ have been enumerated. At that point, one knows for sure that the optimal threshold $\delta^{*}$ must lie in $[\delta_{k},\delta_{k-1})$ and, despite not having all $J$ samples $p_{\mathrm{min}}  \sim \mathrm{PDF}(\Omega^{\prime})$, we do have all of the samples which satisfy $p^{(j)}_{\mathrm{min}} \le \delta_{k-1}$. Since $\delta^{*} < \delta_{k-1}$, those are all that is needed to exactly obtain the optimal corrected significance threshold $\delta^{*}$. On average, the number of samples involved in that computation will be in the order of $\lceil \alpha J \rceil \ll J$.

The pseudocode is shown in Algorithm \ref{alg:incfastwy}. For simplicity, we assume a generic frequent pattern miner which enumerates all patterns in the form of a rooted tree, with children having a support no larger than that of the parent. Due to the removal of limitation (5), adapting the algorithm to deal with the opposite situation, i.e. children having a support not smaller than that of the parent would be trivial. Also, note that the effort required to update $\delta_{k}$ and $\Sigma_{k}$ is absolutely negligible, especially if all values of $\Psi(x)$ for all $x \in [0,N]$ are precomputed at startup.

To summarize, our main contribution is the development of what we believe to be the first practical algorithm to mine statistically significant patterns while controlling the FWER in an optimal sense, i.e., by estimating the empirical null distribution to exactly compensate the dependence structure existing between test statistics. Compared to the current state-of-the-art FastWY algorithm, our proposal improves the following aspects:

\begin{enumerate}
	\item It retrieves the exact set of testable itemsets instead of a surrogate superset.
	\item It uses an incremental search strategy, significantly reducing the frequent pattern mining effort.
	\item It does not need to store the occurrence list of each frequent pattern in memory. 
	\item It does not need to compute the $\lfloor (1-\alpha)J \rfloor$ samples from $p_{\mathrm{min}}  \sim \mathrm{PDF}(\Omega^{\prime})$ which lie in the upper tail of the distribution. This improvement significantly reduces the frequent pattern mining effort further. Besides, the runtime related to the computation of cell counts is also decreased. 
	\item It uses an efficient scheme to share the computation of $p$-values across permutations, reducing the corresponding runtime for that task by a factor of $J$, with $J$ in the order of $10^{4}$. That also avoids caching computations at a further expense of memory usage.
\end{enumerate}

Out of those 5 improvements, 2), 3) and 4) have the greatest overall impact and are all direct consequences of using Properties~\ref{prop:FWER} and~\ref{prop:eval} to derive an incremental search scheme. Improvement 5) is independent of those and roughly halves the runtime in databases for which the computation of cell counts and $p$-values is the main bottleneck. On the contrary, it brings more modest speedups for datasets that have frequent pattern mining as the limiting runtime factor. Finally, improvement 1) is not directly related to computational efficiency, but refines the theoretical background and enhances the applicability of the method.
\section{Experiments}
\label{sec:experiments}

In this section, we demonstrate our five improvements (see Section~\ref{subsec:ouralgorithm}) across a wide range of scenarios with an exhaustive experimental study in two representative data mining problems: {\em significant itemset mining} and {\em significant subgraph mining}.

\subsection{Experimental Setup}

As discussed in Section~\ref{sec:background}, a literal implementation of the FastWY algorithm described in \cite{TeradaIEEE} would lead to impractical runtime requirements for virtually all datasets, leaving us without a comparison partner. Thus we chose to employ the method they actually implemented\footnote{Code available in \url{http://a-terada.github.io/lamp/}.} as a solo comparison partner.

Nonetheless, their code is not efficient enough to provide a fair comparison, mainly due to it being written in Python. As our method has been carefully written in C/C++ with as many fine optimizations as we could include, we reimplemented FastWY from scratch in C/C++, trying to optimize the code as much as that of our own algorithm. As a result, our implementation of FastWY, while being the same algorithm, is about 2 or 3 orders of magnitude faster and reduces the amount of RAM used by one or two orders of magnitude. Thus differences between our method and FastWY which are not due to pure algorithmic considerations but to implementation issues were eliminated.

{\bfseries Itemset Mining Specific Setup:}
As a frequent itemset miner, we chose LCM version 3~\cite{Uno04anefficient} for both of Westfall-Young light and FastWY. LCM has been shown to exhibit state-of-the-art performance in a great number of datasets and won the FIMI'04 frequent itemset mining competition \cite{FIMIDatasets}. Even though
the original implementation of FastWY relies on LCM version 5, we chose LCM version 3 because it is the fastest among all so far according to the author of LCM. For that, we modified the source code, adding the missing features needed to keep track of the occurrence list of each frequent itemset.

All the code was compiled using Intel C++ compiler version 14.0.1 with flags \texttt{-O3 -xavx}. Each instance was executed as a single thread on a cluster computer, whose nodes are equipped with up to 256~GB of RAM and two 2.7~GHz Intel Xeon E5-2697v2 CPUs.

{\bfseries Subgraph Mining Specific Setup:}
In subgraph mining, we employed Gaston~\cite{Nijssen04} as a frequent subgraph miner\footnote{Code available in \url{http://www.liacs.nl/~snijssen/gaston/iccs.html}} because it is reported to be one of the fastest algorithms~\cite{Worlein05}. Both Westfall-Young light and FastWY were integrated into Gaston. All of the methods were written in C++ and compiled with {\ttfamily gcc} 4.9.0 with an option {\ttfamily -03}. We used Mac OS X version 10.9.5 equipped with 32~GB of RAM and a 3.5~GHz Intel Core i7-4771 CPU.

\begin{table*}[t]
\centering
\begin{small}
\caption{Characteristics of itemset mining datasets. $N$ and $n$ are the number of transactions in total and in the minor class, respectively,
$\left\vert E \right\vert$ refers to the number of items, and $\left\vert\left\vert  T \right\vert\right\vert/N$ is the average transaction size.}
\label{tab:dataset_characteristics_itemsets}
	\begin{tabular}{|c|c|c|c|c|c|c|c|c|c|c|c|c|}
		\hline
		\rotatebox[origin=c]{90}{Property} & \rotatebox[origin=c]{90}{TicTacToe} & \rotatebox[origin=c]{90}{Chess} & \rotatebox[origin=c]{90}{Inetads} & \rotatebox[origin=c]{90}{Mushroom} & \rotatebox[origin=c]{90}{Breast cancer} & \rotatebox[origin=c]{90}{Pumsb-star} & \rotatebox[origin=c]{90}{Connect} & \rotatebox[origin=c]{90}{BmsWebview} & \rotatebox[origin=c]{90}{Retail} & \rotatebox[origin=c]{90}{T10I4D100K} & \rotatebox[origin=c]{90}{T40I10D100K} & \rotatebox[origin=c]{90}{Bmspos}      \\
		\hline
		$N$ & $958$ & $3196$ & $3279$ & $8124$ & $12773$ & $49046$ & $67557$ & $77512$ & $88162$ & $100000$ & $100000$ & $515597$\\
		\hline
		$N / n$ & $2.89$ & $-$ & $7.14$ & $2.08$ & $11.31$ & $-$ & $-$ & $-$ & $-$ & $-$ & $-$ & $-$\\
		\hline
		$\left\vert E \right\vert$ & $18$ & $75$ & $1554$ & $117$ & $1129$ & $7117$ & $129$ & $3340$ & $16470$ & $870$ & $942$ & $1657$\\
		\hline
		$\left\vert\left\vert  T \right\vert\right\vert/N$ & $6.93$ & $37.00$ & $12.00$ & $22.00$ & $6.70$ & $50.48$ & $43.00$ & $4.62$ & $10.31$ & $10.10$ & $39.61$    & $6.53$\\
		\hline
	\end{tabular}
\end{small}
\end{table*}

\begin{table*}[t]
\centering
\begin{small}
 \caption{Characteristics of subgraph mining datasets, where $|V|$ and $|E|$ denote the number of vertices and edges, respectively.}
 \label{tab:graphdata}
 \begin{tabular}{|c|c|c|c|c|c|c|c|c|c|c|c|c|}
  \hline
  \rotatebox[origin=c]{90}{Property} & \rotatebox[origin=c]{90}{PTC (MR)} & \rotatebox[origin=c]{90}{PTC (FR)} & \rotatebox[origin=c]{90}{PTC (MM)} & \rotatebox[origin=c]{90}{PTC (FM)} & \rotatebox[origin=c]{90}{MUTAG} & \rotatebox[origin=c]{90}{ENZYMES} & \rotatebox[origin=c]{90}{D\&D} & \rotatebox[origin=c]{90}{NCI1} & \rotatebox[origin=c]{90}{NCI41} & \rotatebox[origin=c]{90}{NCI109} & \rotatebox[origin=c]{90}{NCI167} & \rotatebox[origin=c]{90}{NCI220} \\
  \hline
  $N$ & $584$ & $584$ & $576$ & $563$ & $188$ & $600$ & $1178$ & $4208$ & $27965$ & $4256$ & $80581$ & $900$\\
  \hline
  $N / n$ & $3.23$ & $3.74$ & $3.18$ & $3.15$ & $2.98$ & $2.00$ & $2.42$ & $2.00$ & $17.23$ & $2.00$ & $8.38$ & $3.10$\\
  \hline
  avg.$|V|$ & $31.96$ & $31.96$ & $31.47$ & $31.78$ & $17.93$ & $32.63$ & $284.32$ & $60.12$ & $47.97$ & $59.48$ & $39.70$ & $46.87$\\
  \hline
  avg.$|E|$ & $32.71$ & $32.71$ & $32.18$ & $32.50$ & $39.59$ & $62.14$ & $715.66$ & $62.72$ & $50.15$ & $62.09$ & $41.05$ & $48.52$\\
  \hline
 \end{tabular}
\end{small}
\end{table*}

{\bfseries Itemset Mining Datasets:}
We used 4 labeled datasets: TicTacToe\footnote{\url{https://archive.ics.uci.edu/ml/datasets/Tic-Tac-Toe+Endgame}}, Inetads\footnote{\url{https://archive.ics.uci.edu/ml/datasets/Internet+Advertisements}}, Mushroom, and Breast cancer. The first three are commonly studied datasets taken from the UCI repository: Tic-tac-toe was binarized representing the three possible states (empty, ``x'' or ``o'') of each space in the 3x3 grid with binary indicators; non-binary features and features with missing values were discarded from Inetads; Mushroom was processed as described in \cite{Uno04anefficient}.  Finally, Breast cancer is described in \cite{TeradaIEEE} and used as an example of a challenging dataset, despite the fact that, as we will show, it is far from being the most demanding among our own battery of tests. 

In addition, we used 8 unlabeled datasets from the well-known public benchmark datasets for frequent itemset mining \cite{FIMIDatasets}: Bmspos, BmsWebview, Retail, T10I4D100K, T40-I10D100K, Chess, Connect and Pumsb-star. Since the labels themselves only affect the algorithm via $n$ and $N$ as far as finding the corrected significance threshold $\delta^{*}$ is concerned, we considered two representative cases:
$N / n = 2$ or $10$.
Note that, since $\Psi(x) \approx (N / n)^{-x}$ for $x \ll n$, the more unbalanced the classes are, the larger the resulting testable region will be, resulting in more testable patterns and increased computational demands.
This results in a total of 20 different cases to be tested. We summarize the main properties of each dataset in Table~\ref{tab:dataset_characteristics_itemsets}.

{\bfseries Subgraph Mining Datasets:}
We used 12 labeled graph datasets: four PTC (Predictive Toxicology Challenge) datasets\footnote{\url{http://www.predictive-toxicology.org/ptc/}}, MUTAG, ENZYMES, D\&D\footnote{MUTAG, ENZYMES, and D\&D are obtained from \url{http://mlcb.is.tuebingen.mpg.de/Mitarbeiter/Nino/Graphkernels/data.zip}}, and four NCI (National Cancer Institute) datasets\footnote{\url{https://pubchem.ncbi.nlm.nih.gov/}}, where ENZYMES and D\&D are proteins and others are chemical compounds.
These datasets are popular benchmarks and have been frequently used in previous studies (e.g.~\cite{Li12,Shervashidze11}).
Graph nodes are labeled in all datasets and edges are also labeled except for ENZYMES and D\&D.
In the four PTC datasets, graphs labeled as CE, SE, or P were treated as positive and those of NE or N as negative, the same setting as in~\cite{Kong10}. Properties of these datasets are summarized in Table~\ref{tab:graphdata}.

Note that the number of nodes in subgraphs is bounded under $15$ in NCI1, NCI109, and NCI220, $10$ in MUTAG, NCI41, and NCI167, and $8$ in ENZYMES so that the comparison partner, FastWY, can finish in a reasonable time to check its peak memory consumption. For example, in ENZYMES with the maximum subgraph size $10$, our method takes $3.6$ hours while FastWY did not stop after two weeks.
In D\&D and the four PTC datasets, the size of subgraphs is unlimited.

\subsection{Results}

\begin{figure}[t]
\centering
\begin{subfigure}{\linewidth}
  \centering
  \includegraphics[width=.99\linewidth]{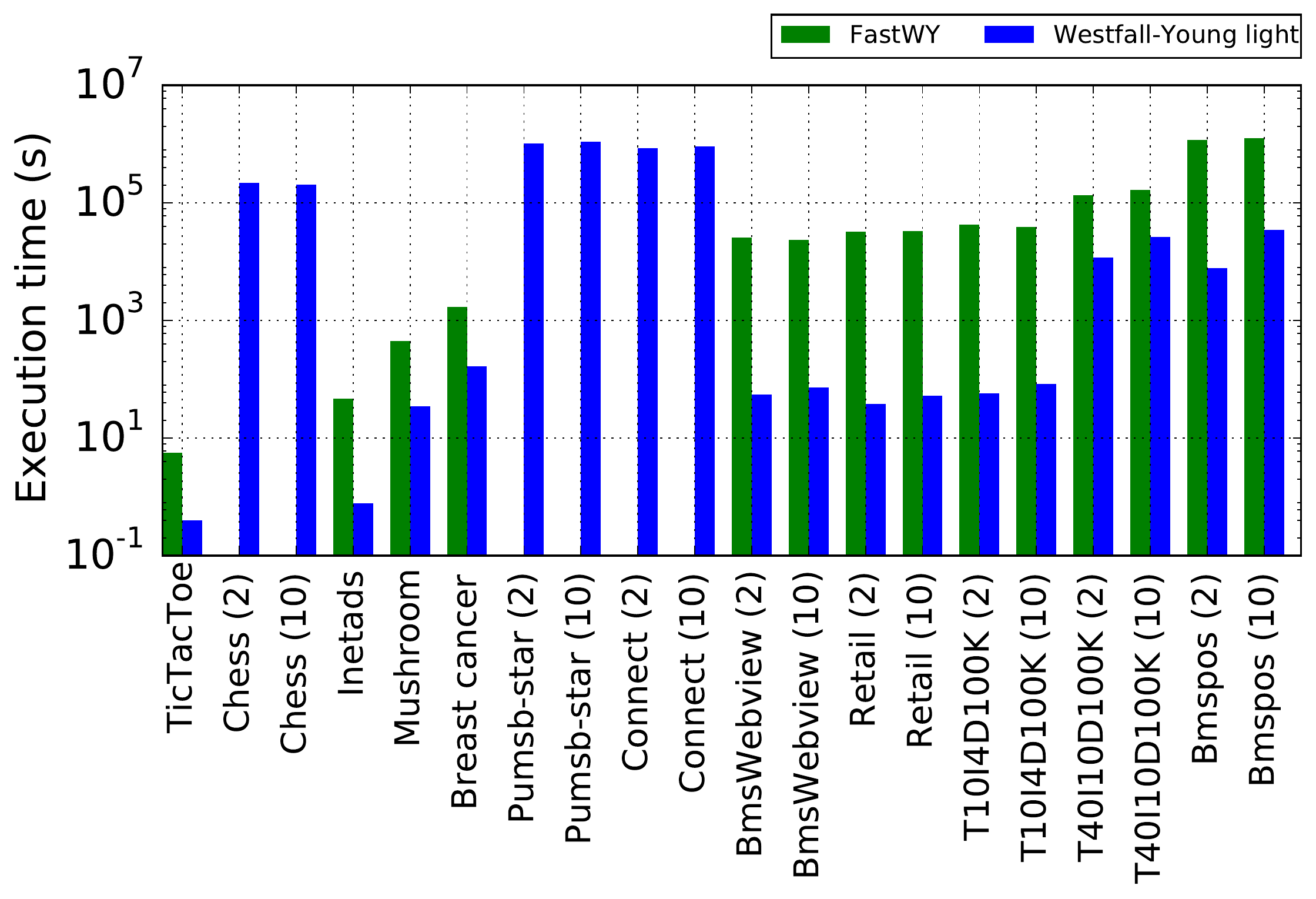}
  \caption{Runtime}
  \label{fig:runtime_comparison_itemsets}
\end{subfigure}%
\newline\vspace*{5pt}
\begin{subfigure}{\linewidth}
  \centering
  \includegraphics[width=.99\linewidth]{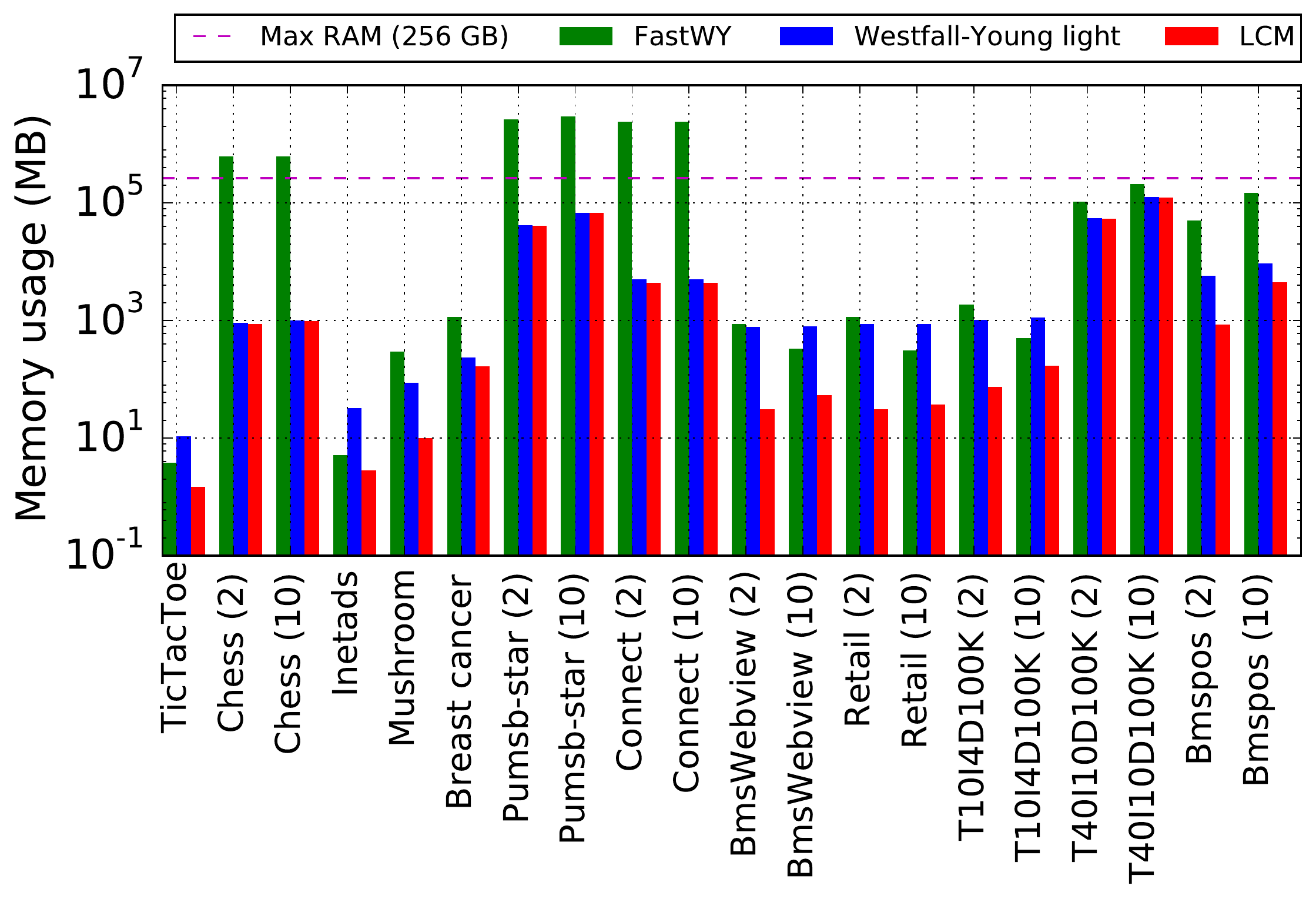}
  \caption{Peak memory usage}
  \label{fig:memory_comparison_itemsets}
\end{subfigure}
\caption{Performance comparison between Westfall-Young light and FastWY in itemset mining with 10,000 permutations. Missing runtime points for FastWY correspond to those cases for which the algorithm crashed due to excessive memory requirements. We include the memory usage due to LCM, which corresponds to the bare minimum amount of memory needed to complete these tasks. Numbers attached to the names of datasets denote the class ratio $N / n$.}
\label{fig:comparison_itemsets}
\end{figure}

\begin{figure}[t]
\centering
\begin{subfigure}{\linewidth}
  \centering
  \includegraphics[width=.99\linewidth]{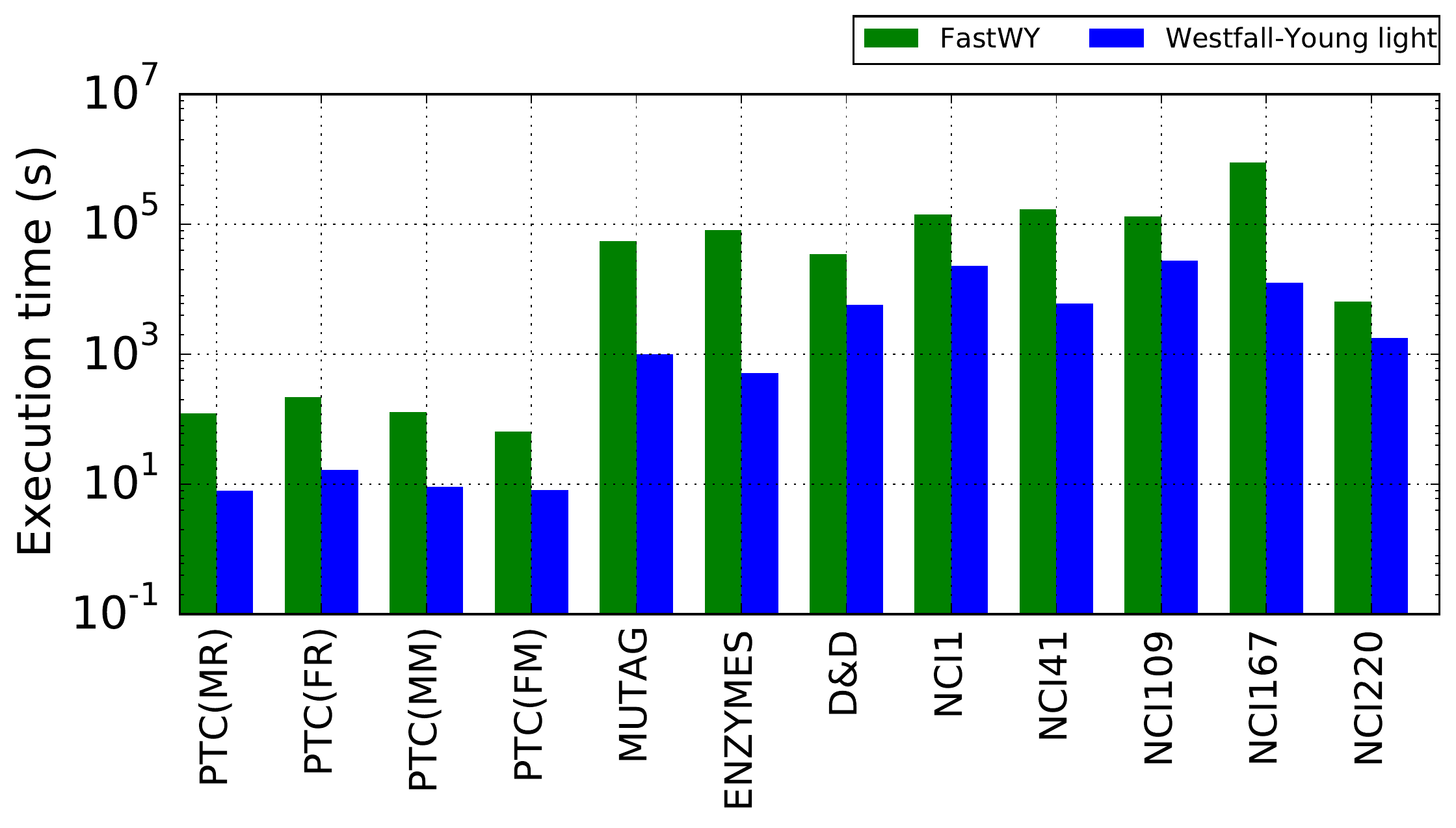}
  \caption{Runtime}
  \label{fig:runtime_comparison_fsm}
\end{subfigure}%
\newline\vspace*{5pt}
\begin{subfigure}{\linewidth}
  \centering
  \includegraphics[width=.99\linewidth]{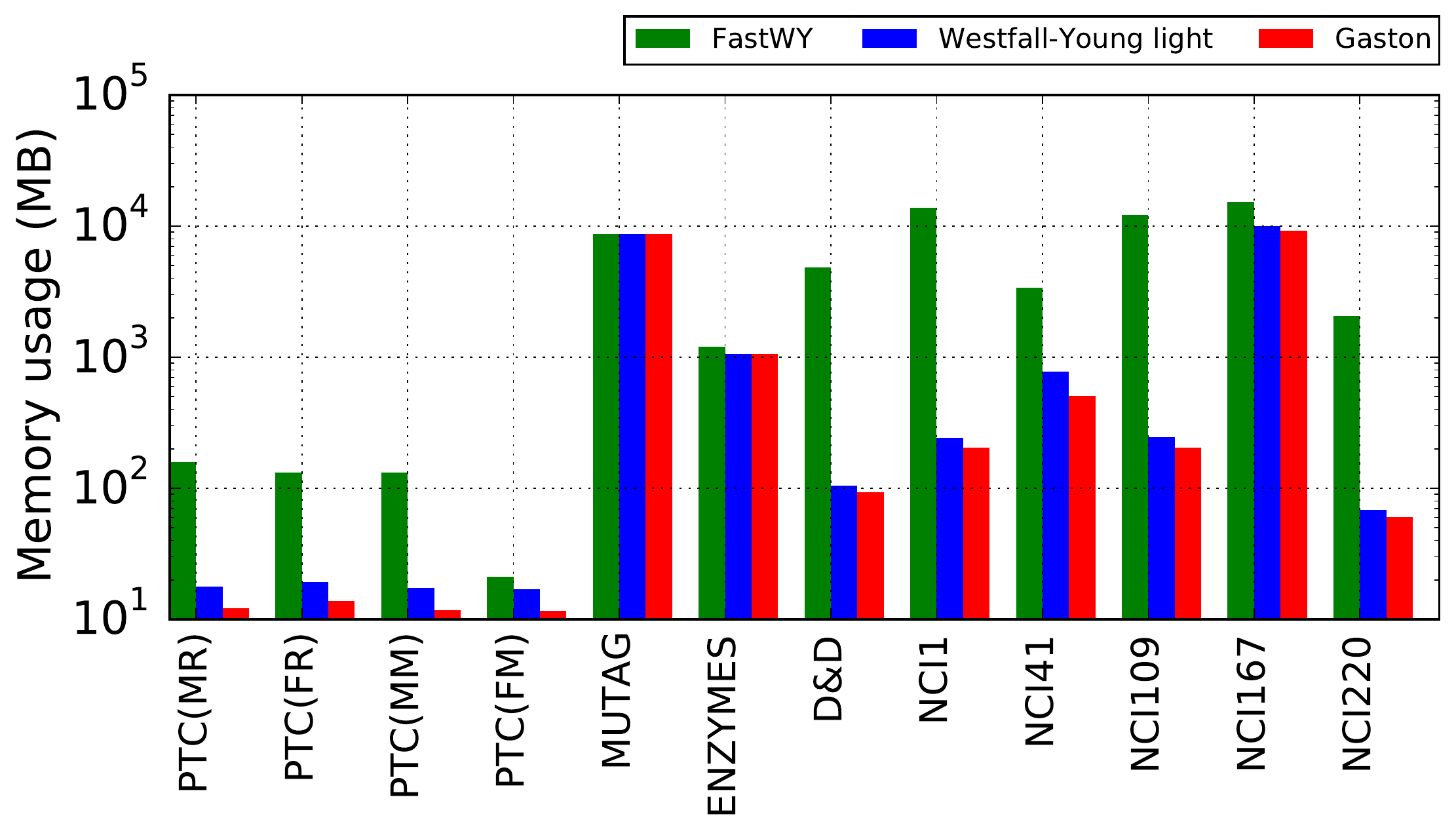}
  \caption{Peak memory usage}
  \label{fig:memory_comparison_fsm}
\end{subfigure}
\caption{Performance comparison between Westfall-Young light and FastWY in subgraph mining with 10,000 permutations. The memory usage due to Gaston is included.}
\label{fig:comparison_fsm}
\end{figure}

\subsubsection*{Runtime and memory usage}

As the main result, we compare the runtime and memory usage of our method Westfall-Young light and the comparison partner FastWY for all 20 itemset mining cases and 12 subgraph mining datasets. In both algorithms, the number $J$ of permutations is the only parameter. We set $J=10^{4}$ for a reason which shall be discussed later. Recall that, as discussed in Section~\ref{sec:introduction}, the target FWER $\alpha$ is not a parameter, but a user requirement. By far, the most standard choice across different scientific disciplines is $\alpha=0.05$, which is the option we used in this experiment.  

The results for significant itemset mining are summarized in Figure~\ref{fig:comparison_itemsets}. We can see that our algorithm reduces the runtime by 3 orders of magnitude in 6 out of 20 cases, by 2 orders of magnitude in 3 cases and by 1 order of magnitude in 5 cases. Even more importantly, for the remaining 6 cases corresponding to datasets Chess, Pumsb-star and Connect, our comparison partner crashed due to excessive memory requirements. 

Indeed, as far as memory usage is concerned, we see two very different situations: (1) in 70~\% of the cases, both methods use approximately the same amount of memory (up to the order of magnitude) but; (2) in the remaining 30~\%, the peak memory usage of FastWY soars up to the point in which the algorithm simply crashes. The actual memory usage of FastWY for the cases in which it crashed is in fact severely underestimated in Figure~\ref{fig:memory_comparison_itemsets}; the numbers we plotted are simply a generous lower bound on what the actual memory usage would have been, obtained with a technique we will describe shortly. Note also that, for many of the datasets for which both methods have about the same peak RAM usage, most of that memory is used by the frequent itemset miner LCM. The memory used by LCM is a constant offset for both methods, which explains why the memory performance is so similar for many datasets. More importantly, one can see that the memory overhead of our algorithm scales very gently across datasets, and the overhead is negligible for about the half of the cases. On the other hand, FastWY shows very poor memory scaling; as soon as the databases get large and dense (see Chess, Connect or Pumsb-star), the algorithm completely breaks down.

Another interesting observation is that, despite the fact that Westfall-Young light improves the runtime of FastWY by only one order of magnitude in 25\% of the cases, those are precisely the datasets for which the total runtime is quite small, below half an hour, except for T40I10D100K. In other words, we claim that the more computationally-demanding the transaction database is, the larger the runtime gap between Westfall-Young light and FastWY gets. That, together with the ill-conditioned scaling of peak memory usage that FastWY exhibits, makes our proposal a superior choice for large-scale significant pattern mining.

We can confirm the same trend in subgraph mining from results summarized in Figure~\ref{fig:comparison_fsm}.
Our method Westfall-Young light is orders of magnitude faster than FastWY across all graph datasets. Moreover, Westfall-Young light reduces the peak memory usage by one to two orders of magnitude in 8 out of 12 cases.
Although Westfall-Young light is always faster than FastWY, the runtime gap is smaller than in the itemset mining case. The reason is that, in subgraph mining, most of computation is devoted to the mining process, which again is a constant offset for both methods. 
In terms of memory usage, Westfall-Young light has a negligible overhead for all subgraph datasets; virtually all memory used by our algorithm is in fact the memory needed by Gaston to carry out frequent subgraph mining. That is in sharp contrast with FastWY, which often requires significantly more memory to store all occurrence lists.

\subsubsection*{Complexity analysis}

At this point, one should ponder what causes the wide spread in runtime between different datasets. By simply comparing Figures~\ref{fig:comparison_itemsets} and~\ref{fig:comparison_fsm} to Tables~\ref{tab:dataset_characteristics_itemsets} and~\ref{tab:graphdata}, one can see that the magnitudes listed in the tables do not correlate with runtime in a clear way. In this section, we will derive the best predictor for runtime given the characteristics of a dataset. 

\begin{figure}[t]
\centering
\includegraphics[width=.7\linewidth]{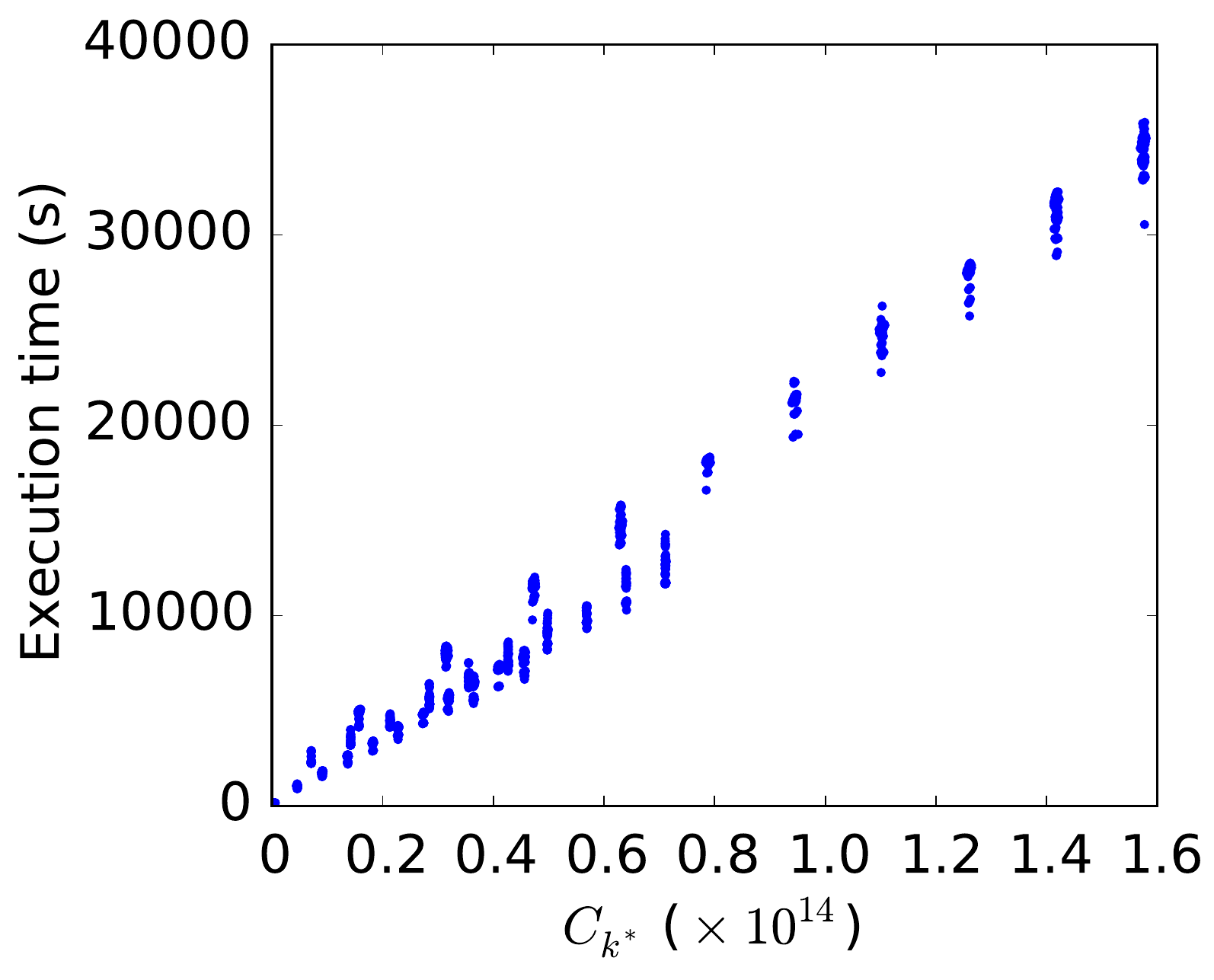}
\caption{Execution time (s) versus $C_{k^{*}}$ for itemset mining datasets with $10^{2}$ repetitions of each case.
Both for computational and visualization reasons, the most demanding datasets have been left out from the picture, but can be readily checked to follow the same quasi-linear trend. }
\label{fig:datasetcost_scatter_itemsets}
\end{figure}

Let us define $c(x) = \sum_{i=1}^{D}$ ${\mathbbm{1}[x_{i} = x]}$, the number of patterns which have support $x$ in the database. We call the quantity $C_{k} = \sum_{x \in \Sigma_{k}}{x c(x)}$ the \emph{total dataset cost in region $\Sigma_{k}$}, which is the sum of the support of every single pattern which is testable in region $\Sigma_{k}$. As shown in Figure~\ref{fig:datasetcost_scatter_itemsets}, $C_{k^{*}}$ with $k^{*}$ being the value of $k$ when Algorithm \ref{alg:incfastwy} terminates, turns out to be clearly correlated with the total runtime. We confirmed the same trend in subgraph mining. That is hardly surprising since the total effort to compute cell counts is $O(JC_{k^{*}})$ and, on the other hand, the mining effort scales with the non-weighted dataset cost $\widetilde{C}_{k} = \sum_{x \in \Sigma_{k}}{c(x)}$, which is naturally related to the weighted-counterpart $C_{k}$. All other factors contributing to runtime are negligible for Westfall-Young light, as discussed in Section~\ref{sec:proposal}.

$C_{k}$ is not only a good predictor for the runtime but, also, it is an almost exact proxy for the memory overhead of our comparison partner, FastWY. Since the method in \cite{TeradaIEEE} relies on storing the occurrence list of every frequent pattern in memory, assuming that each occurrence is stored as a 4-byte integer, the amount of RAM required for that purpose would be roughly $4C_{k_{\mathrm{min}}}$ bytes. There $k_{\mathrm{min}}$ is the minimum value that $k^{*}$ takes among all $J$ permutations in FastWY. On the contrary, the memory overhead of our algorithm is completely unrelated to $C_{k}$, being simply $NJ$ bytes if whole chars are used to store each binary entry (as we did) or $NJ/8$ bytes if the matrix is packed. 

As seen in the experiments, for small and medium-size databases, $NJ$ and $4C_{k_{\mathrm{min}}}$ appear to be roughly of the same order of magnitude. Both methods exhibiting similar memory requirements in 70\% of the itemset and 33\% of the subgraph mining cases is a consequence of that. Moreover, as we indicated previously, the peak memory usage is actually dominated by the requirements of the frequent pattern miner for many of those datasets. In more demanding datasets, $NJ$ scales gently whereas $C_{k}$ exhibits a combinatorial explosion. This is why FastWY cannot scale to large databases, leaving Westfall-Young light as the first method that can deal with massive, challenging databases.

\begin{figure}[t]
\centering
\begin{subfigure}{\linewidth}
  \centering
  \includegraphics[width=.95\linewidth]{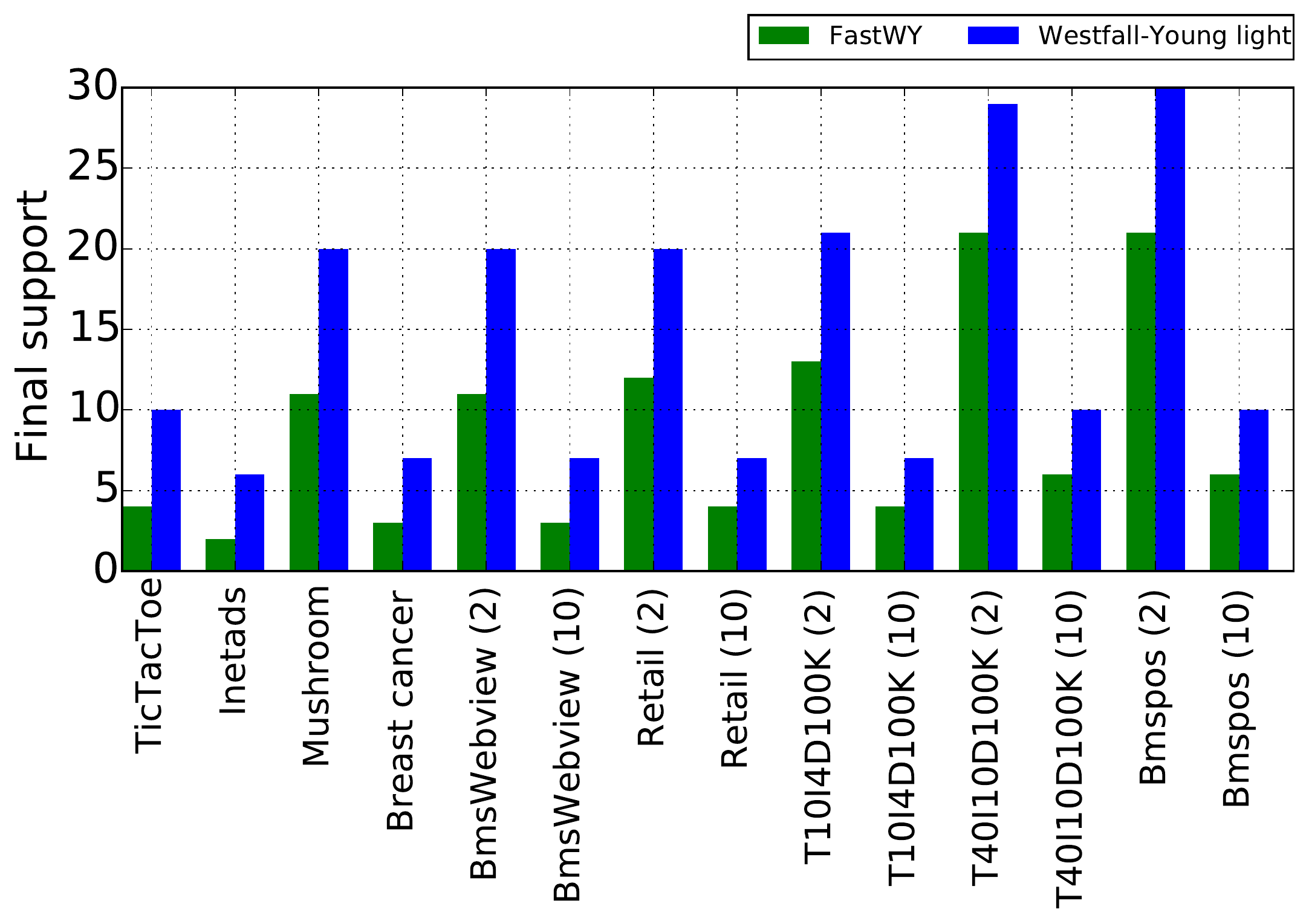}
  \caption{Itemset mining.}
\end{subfigure}%
\newline\vspace*{5pt}
\begin{subfigure}{\linewidth}
  \centering
  \includegraphics[width=.95\linewidth]{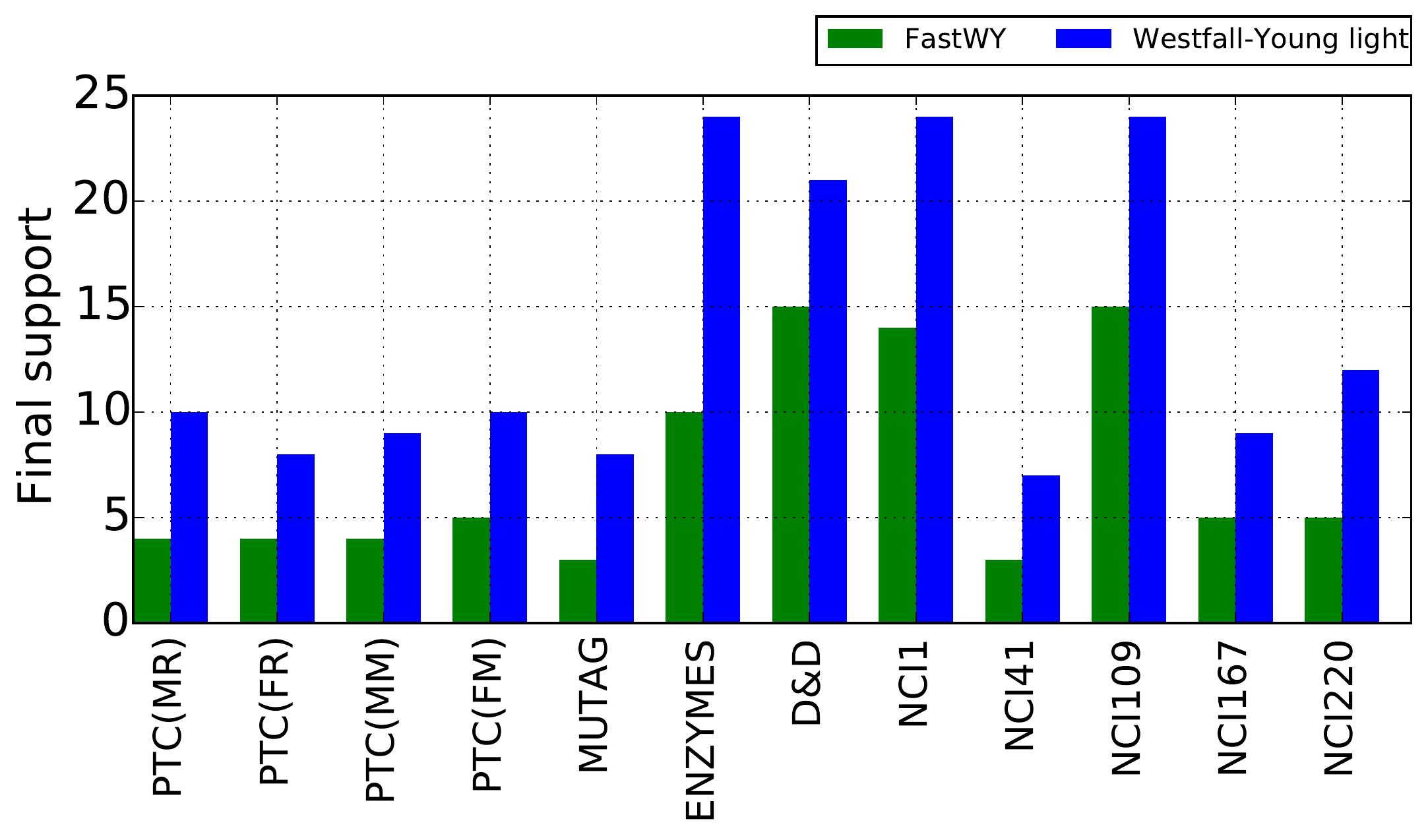}
  \caption{Subgraph mining.}
\end{subfigure}
\caption{Final support when using Westfall-Young light, $\sigma^{k^{*}}_{l}$, versus the worst case support in FastWY which determines its memory and runtime requirements, $\sigma^{k_{\mathrm{min}}}_{l}$.}
\label{fig:final_support_itemsets}
\end{figure}

Another crucial fact is that, because FastWY treats each permutation independently and requires to exactly generate all $J$ samples from $\mathrm{PDF}(\Omega^{\prime})$, it needs to mine as many patterns as required by the worst case across all $J$ permutations.
In other words, the runtime depends on $k_{\mathrm{min}}$, which is the worst-case realisation of $k^{*}$ across $J$ permutations. In practice, $k_{\mathrm{min}}$ can be much smaller than $k^{*}$ or, equivalently, the minimum support $\sigma^{k_{\mathrm{min}}}_{l}$ can be much smaller than $\sigma^{k^{*}}_{l}$. In other words, FastWY consistently needs to mine frequent patterns with much lower supports than Westfall-Young light and to store all their occurrence lists in RAM. Actual numbers are plotted in Figure~\ref{fig:final_support_itemsets}. 

This effect is particularly harmful for FastWY as most databases obey a power-law distribution in $c(x)$, making patterns with low supports much more abundant in the database that those with large supports. Consequently, even small gaps between $\sigma^{k_{\mathrm{min}}}_{l}$ and $\sigma^{k^{*}}_{l}$ can result in large runtime and storage overheads. In fact, also a direct result of the power-law distribution affecting $c(x)$, the smaller $\sigma^{k^{*}}_{l}$ the more impact the gap between $\sigma^{k_{\mathrm{min}}}_{l}$ and $\sigma^{k^{*}}_{l}$ will have in the computational demands. This is particularly relevant for unbalanced datasets, i.e. those with large $N/n$, as they typically have low values of $\sigma^{k^{*}}_{l}$. To summarize, {\em the computational complexity of FastWY is lower bounded from the worst case scenario out of $J$ permutations, while Westfall-Young light completely bypasses this ill-posed dependence by generating only $\lceil \alpha J\rceil$ samples from $\mathrm{PDF}(\Omega^{\prime})$ exactly}.


As a side remark, the peak memory usage results for FastWY of 3 datasets (Chess, Pumsb-star, and Connect) shown in Figure \ref{fig:memory_comparison_itemsets} were obtained as $4C_{k^{*}}$ (expressed in MB). As we said, that is a fairly generous lower bound on the actual memory usage for two reasons: (1) it does not consider other contributions to the peak memory usage, such as the RAM needed to run the frequent pattern miner and; (2) it is obtained from $C_{k^{*}}$ and not $C_{k_{\mathrm{min}}}$, which can be significantly larger as we just discussed. Nevertheless, given the impossibility to actually run the algorithm without crashing, it is the best approximation we could achieve.

\subsubsection*{Choosing the number of permutations $J$ and effect of addressing the dependence between test statistics on the resulting statistical power}

As far as parameters are concerned, we must only deal with the number $J$ of permutations. Intuitively, the trade-off involved when setting $J$ is clear: the larger $J$, the more precise the estimation of $\delta^{*}$ will be at the expense of increased runtime. In problems for which the frequent pattern mining effort is the dominant runtime factor, such as subgraph mining, the execution time increases sublinearly with $J$. On the other hand, in problems for which computing cell counts is the main bottleneck, like most itemset mining examples, the increase will be almost exactly linear.

To illustrate the effect of changes in the number $J$ of permutations, we executed Westfall-Young light for 10 different values of $J$ between $J=10^{3}$ and $J=10^{4}$ in steps of $\Delta J=10^{3}$. For each pair (dataset, $J$) we repeat the execution 100 times and show the median empirical $\mathrm{FWER}$ as a function of $J$, along with the corresponding 5\%--95\% confidence interval. Furthermore, we also show the resulting $\mathrm{FWER}$ obtained by using the corrected significance threshold obtained with LAMP, $\delta_{\mathrm{LAMP}}$~\cite{TeradaPNAS}, which is the first prominent method that controls the FWER in pattern mining (see Section~\ref{sec:related_work} for more detail). Note that the FWER with the Bonferroni factor must be always smaller than that with LAMP.

The reason why we chose the empirical $\mathrm{FWER}$ as the quantity of interest is that it is the best proxy for the resulting statistical power of the method. Ideally, an algorithm for which $\mathrm{FWER}=\alpha$ would be optimal among all statistical testing procedures controlling the FWER at level $\alpha$. On the other hand, if the resulting empirical $\mathrm{FWER}$ satisfies $\mathrm{FWER} < \alpha$, there is a loss of power due to the scheme being too conservative. When dealing with discrete test statistics, as in the case of this paper, achieving $\mathrm{FWER}=\alpha$ might not always be possible yet the reasoning remains the same: the closer $\mathrm{FWER}$ is to $\alpha$ (without ever being bigger), the better. To summarize, the purpose of this experiment is twofold: (1) we can see the effect that correcting for dependence via permutation-testing has on the resulting statistical power and (2) we can show that increasing $J$ beyond a certain value has a very small effect in the resulting empirical $\mathrm{FWER}$ and its spread. 

\begin{figure}[t]
\centering
\begin{subfigure}{.5\linewidth}
  \centering
  \includegraphics[width=.99\linewidth]{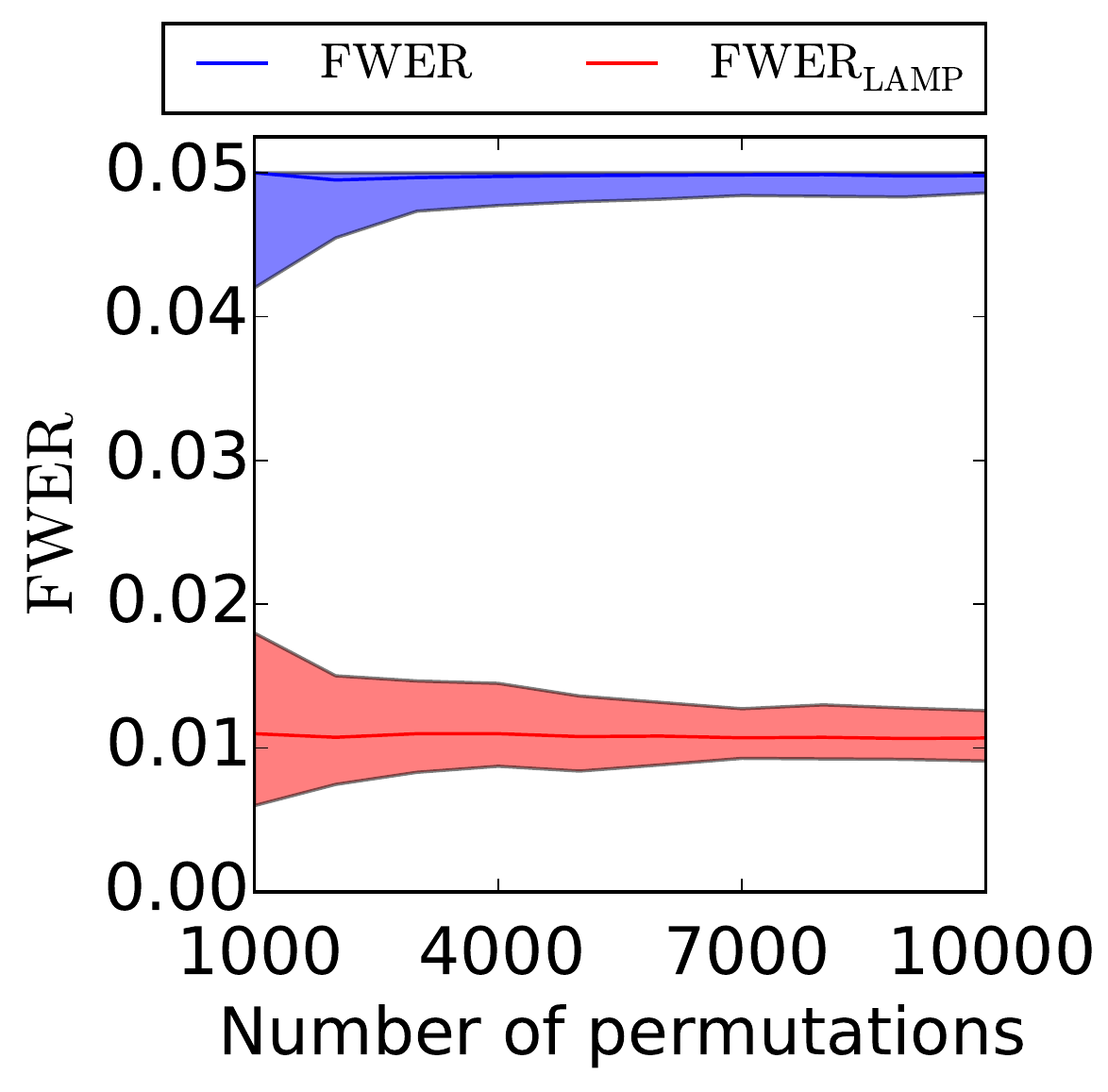}
  \caption{BmsWebview ($N / n=2$)}
  \label{fig:fwer_itemsets1}
\end{subfigure}%
\begin{subfigure}{.5\linewidth}
  \centering
  \includegraphics[width=.99\linewidth]{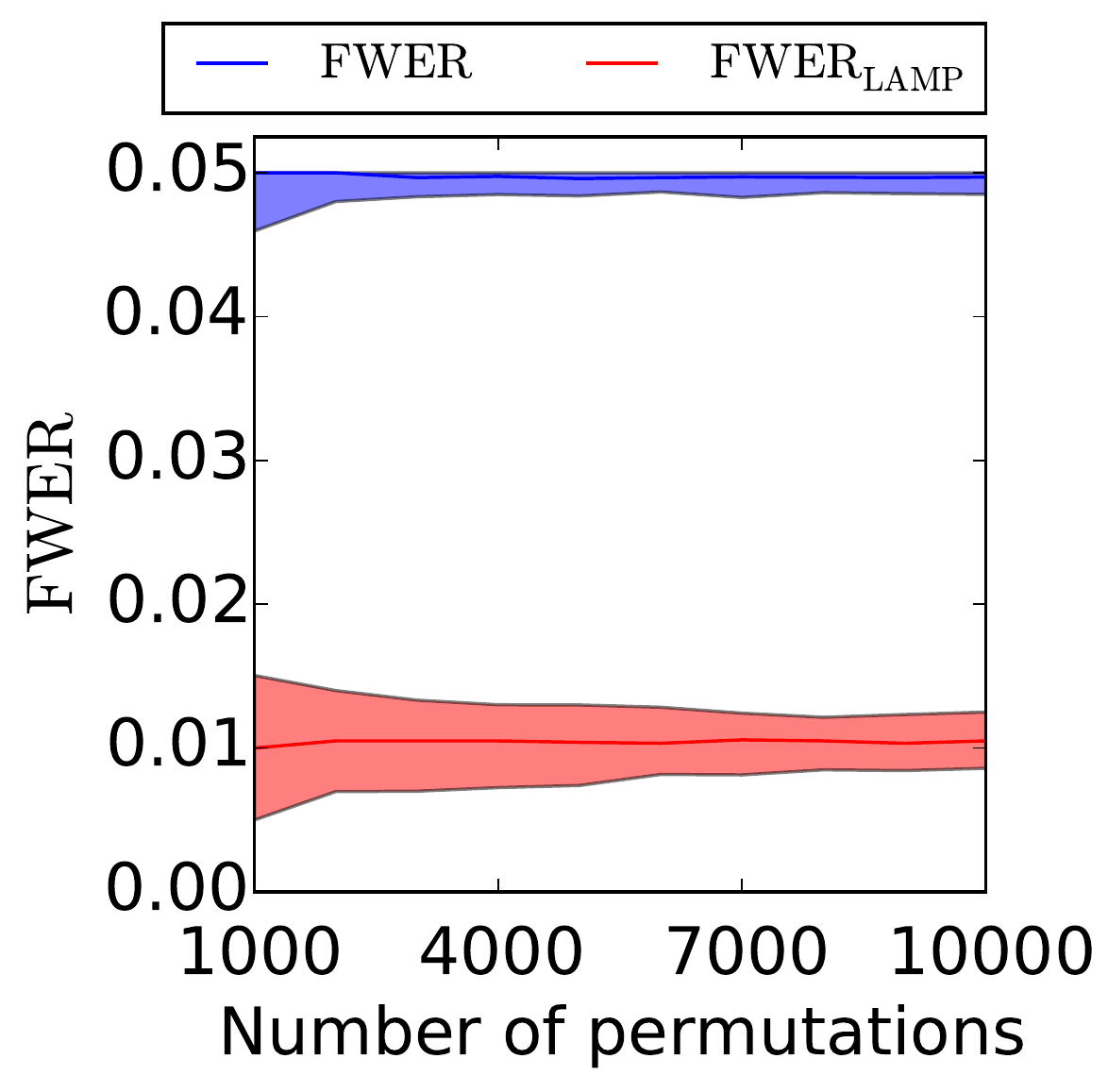}
  \caption{T40I10D100K ($N / n=2$)}
  \label{fig:fwer_itemsets2}
\end{subfigure}
\caption{Empirical $\mathrm{FWER}$ versus $J$ for two representative itemset mining databases.}
\label{fig:fwer_itemsets}
\end{figure}

\begin{figure}[t]
\centering
\begin{subfigure}{.5\linewidth}
  \centering
  \includegraphics[width=.99\linewidth]{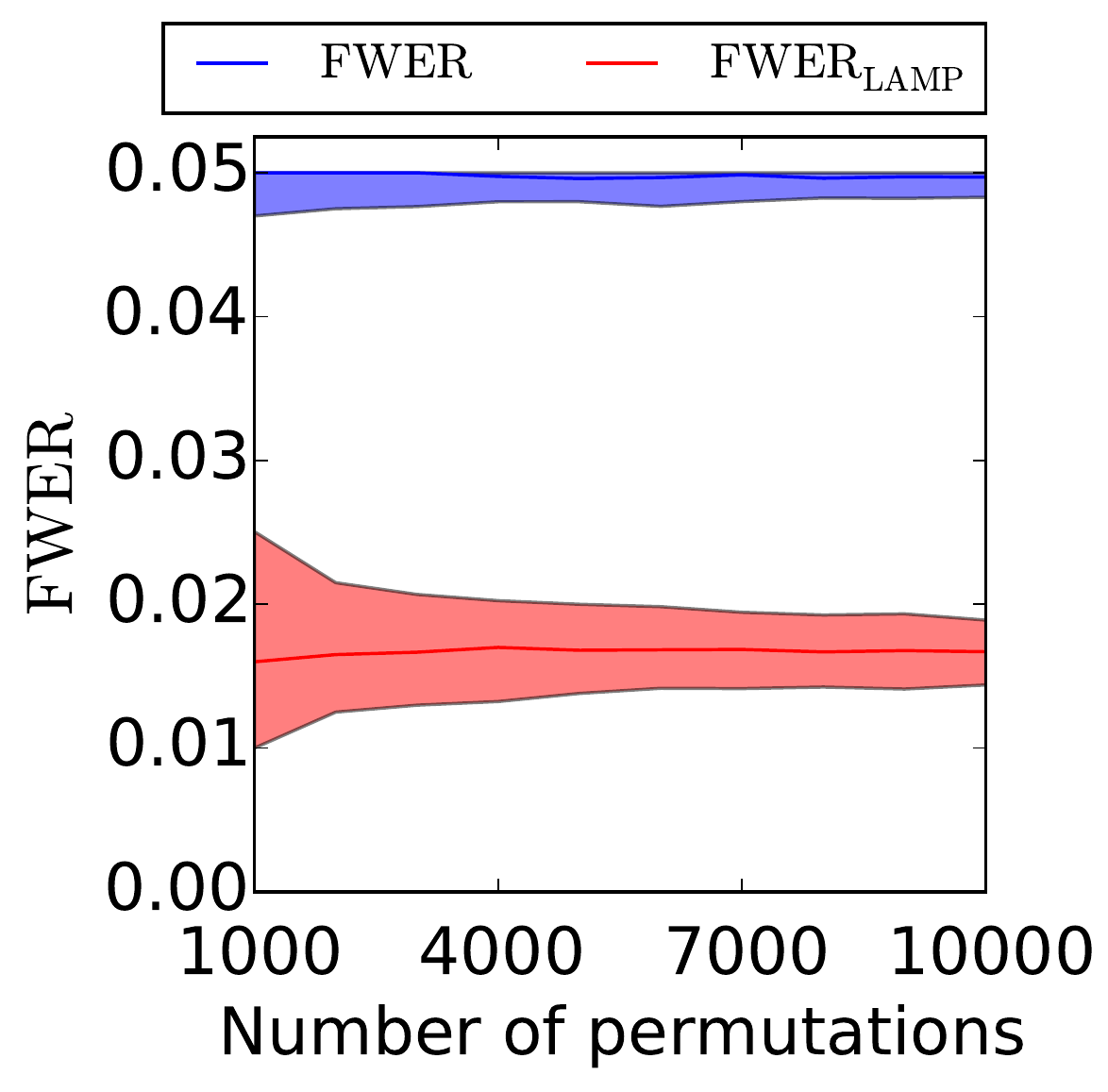}
  \caption{ENZYMES.}
  \label{fig:fwer_subgraph1}
\end{subfigure}%
\begin{subfigure}{.5\linewidth}
  \centering
  \includegraphics[width=.99\linewidth]{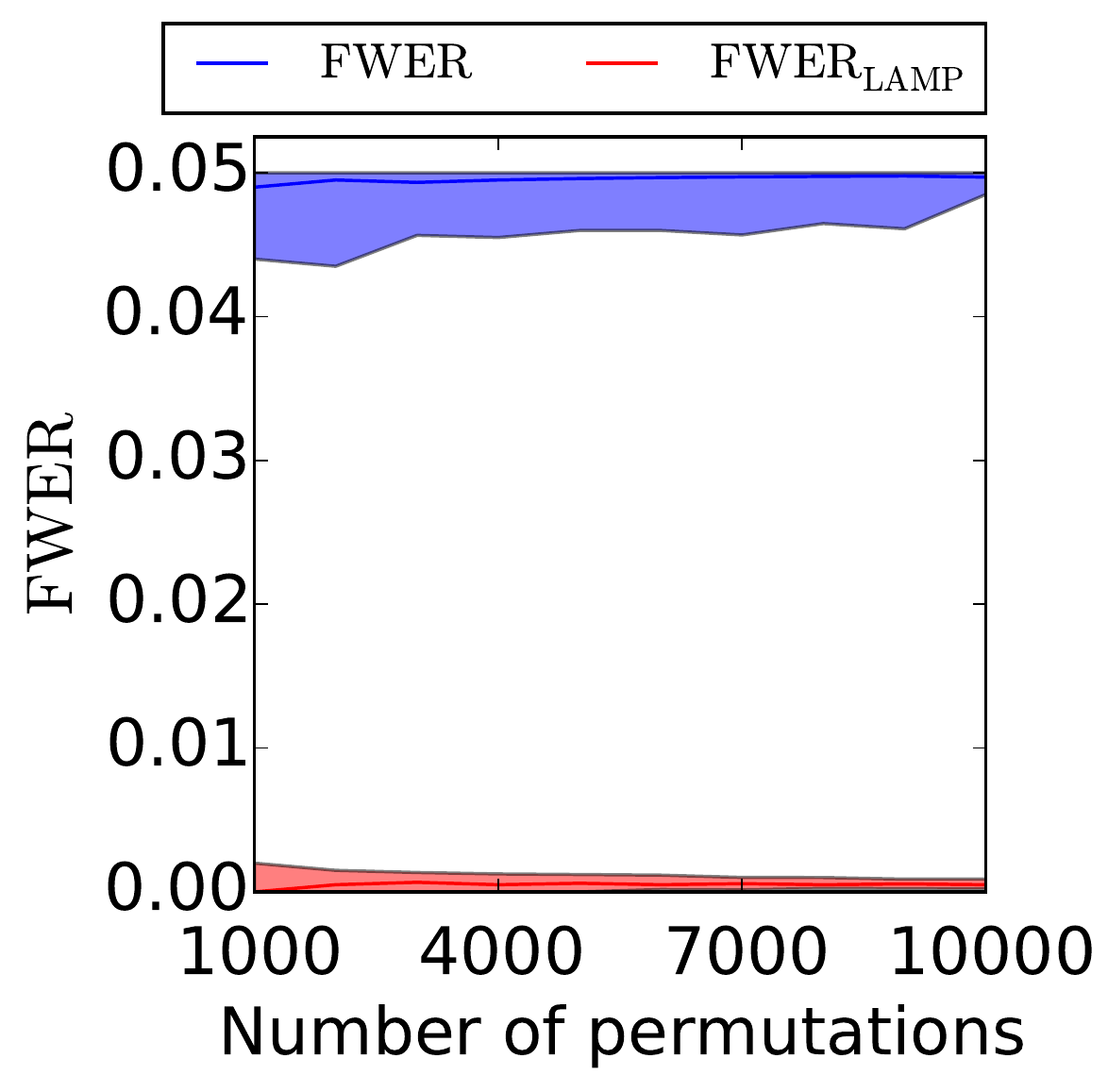}
  \caption{NCI220.}
  \label{fig:fwer_subgraph2}
\end{subfigure}
\caption{Empirical $\mathrm{FWER}$ versus $J$ for two representative subgraph mining databases.}
\label{fig:fwer_subgraphs}
\end{figure}

We depict the results for four sample datasets due to space considerations: two datasets in itemset mining (BmsWebview and T40I10D100K) in Figure~\ref{fig:fwer_itemsets} and two datasets in subgraph mining (ENZYMES and NCI220) in Figure~\ref{fig:fwer_subgraphs}. As it can be seen, the median $\mathrm{FWER}$ appears to be fairly stable to changes in $J$ and, more importantly, the spread of the empirical $\mathrm{FWER}$ saturates at about $J=10^{4}$. Therefore, we believe $J=10^{4}$ to be the safest parameter choice. Moreover, for runtime-limited scenarios for which cell count computations are the main bottleneck, using $J=10^{3}$ might still lead to good performance while reducing the runtime by about one order of magnitude.

Finally, our results also clearly demonstrate that LAMP is a severely over conservative algorithm. It tends to yield an empirical $\mathrm{FWER}$ which oscillates between $\alpha / 2$ and $\alpha / 100$ depending on the dataset.
Even just halving the target FWER can have drastic consequences in the resulting statistical power: when dealing with massive multiple-hypothesis testing, as in the case of significant pattern mining, picking up subtle signals is the main objective. A large amount of significant patterns might be lost as a consequence of over-controlling the $\mathrm{FWER}$.
\section{Related Work}
\label{sec:related_work}

We briefly describe other existing methods for mining statistically significant patterns. As discussed in the introduction, the overwhelming majority of papers that propose algorithms to find significant patterns do not address multiple testing. Thus the soundness of their approaches is not reliable since they do not provide provable guarantees on the probability or proportion of false discoveries being made by the algorithm. An alternative approach~\cite{Webb07} is to limit the maximum size of the patterns being enumerated to reduce the overall number of tests, thus reducing the loss of statistical power due to multiple-testing. However, since the number of tests (patterns) scales combinatorially, that method can deal with only very small patterns before losing all statistical power. In the following we review only those algorithms which rigorously correct for multiple testing without imposing restrictive limits on the pattern size.

The concept of minimum attainable $p$-value for discrete test statistics and its usefulness for multiple hypothesis testing was first noticed by Tarone~\cite{Tarone}. He realized that, given a tentative corrected significance threshold $\delta$ and just based on the margins $x$, $n$, and $N$ of the $2 \times 2$ contingency table, one can conclude that all patterns for which $\Psi(x_{i}) > \delta$ have no chance of being significant and can be pruned from the search space. Even more importantly, because those patterns can never be significant, they do not need to be included in the Bonferroni correction factor.

Precisely, let $m(\delta)=\left\vert \mathcal{I}_{T}(\delta) \right\vert$. Then, Tarone proved that $\mathrm{FWER} \le \delta m(\delta)$ and that an improved corrected significance threshold can be found as $\delta^{*} = \max\set{\delta | \delta m(\delta) \le \alpha}$. Since $m(\delta) \ll D$ in most real-world datasets, this improvement can bring a drastic increase in statistical power.

This improved Bonferroni correction for discrete data requires computing $\Psi(x_{i})$ for each pattern to evaluate $m(\delta)$ as many times as needed in a linear search. Needless to say, in a data mining context, that approach is unfeasible as it would require mining and computing the support of every single pattern in the dataset. The computational unfeasibility of the original method resulted in it being largely ignored by the data mining community until the 2013 breakthrough paper by Terada et al.~\cite{TeradaPNAS}. There, the authors propose the Limitless-Arity Multiple-testing Procedure (LAMP), a branch-and-bound algorithm that uses a frequent itemset mining algorithm to efficiently apply Tarone's method to data mining problems of the form considered in this paper.

On the algorithmic side, a decremental search strategy was adopted in LAMP, same as FastWY. They let $\sigma=n$ at initialization and solve a sequence of frequent pattern mining problems, using the miner as a black box to evaluate $\hat{m}(\sigma) = |\hat{\mathcal{I}}_{T}(\sigma(\delta)) |$ for each $\sigma$ until the condition $\hat{\Psi}(\sigma) \hat{m}(\sigma) \le \alpha$ is violated for some value of $\sigma$. When that occurs, one can conclude that the optimal $\sigma^{*}$ satisfies $\sigma^{*} = \sigma+1$.

LAMP can be seen as the first successful instance of multi-ple-testing correction applied to significant pattern mining, as it can discover significant itemsets of arbitrary size while upper bounding the FWER. Moreover, it has a reasonable runtime and storage complexity, at least in small and mid-sized datasets. In order to tackle larger problems, two studies~\cite{Minato} and~\cite{Sugiyama15SDM} simultaneously proposed changing the search strategy to an incremental scheme with some form of early stopping; the first in the context of subgraph mining and the latter to deal with itemset mining problems. The basic idea is to initialize $\sigma=1$ and iteratively increase $\sigma$ every time the condition $\hat{\Psi}(\sigma) \hat{m}(\sigma) \le \alpha$ is found to be violated. The frequent pattern mining algorithm does not need to be restarted every time $\sigma$ changes, making the whole process efficient. Empirically, this new strategy has been shown to bring a large runtime reduction in both scenarios. 

Nevertheless, neither the conceptual method described in \cite{Tarone}, its first practical implementation in \cite{TeradaPNAS} nor the runtime-optimized version in \cite{Minato} address the dependence structure existing between test statistics. Consequently, they are strictly suboptimal testing procedures which lose a large fraction of statistical power by overestimating the FWER (see Figure~\ref{fig:fwer_itemsets}). In~\cite{Sugiyama15SDM}, we tried to alleviate that problem by exploiting the concept of \emph{effective number of tests}, estimated by using classical permutation-testing only on the subset of testable subgraphs.  Since the method in~\cite{Sugiyama15SDM} does not use algorithmic tricks to apply permutation-testing, its computational feasibility relies on the set of testable subgraphs being small enough. Nevertheless, that is often the case in practice. All-in-all, that idea constitutes a step towards addressing dependence, but it is still a heuristic, strictly suboptimal approach, which controls the FWER more than needed.

To the best of our knowledge, the FastWY algorithm in \cite{TeradaIEEE} is the only attempt so far to optimally take the dependence between test statistics in pattern mining into account. However, as discussed in Section~\ref{sec:background}, FastWY is based on a decremental search scheme instead of the more efficient incremental counterpart. Moreover, as FastWY has to repeat the decremental search $J \approx 10^{4}$ times, including the pattern mining effort, the performance gap between decremental and incremental search is expected to be even more dramatic. Furthermore, as we have shown in Section~\ref{sec:experiments}, that runtime gap can still be as high as 3 orders of magnitude even when very large amounts of RAM are traded-off for speed.

\section{Conclusions}
\label{sec:conclusions}


In this paper, we have described a novel algorithm, called {\em Westfall-Young light}, for mining statistically significant patterns which allows the user to adjust exactly the probability of having false discoveries. It estimates the null distribution of the test statistics via Westfall-Young permutations, and succeeds to overcome the massive computational cost of permutation testing in large databases by exploiting a set of computational tricks.

Empirically, our Westfall-Young light algorithm drastically improves upon the state-of-the-art: The runtime decreases by up to three orders of magnitude and the peak memory usage by one up to two orders of magnitude in several itemset and subgraph mining benchmarks. Moreover, we also show that the peak memory usage of Westfall-Young light scales gently with the complexity of the database. In contrast, the peak memory usage of the state-of-the-art algorithm soars as the databases get large and dense, thus breaking down in large-scale problems.

Several interesting challenges still remain to be addressed: In domains such as computational biology, there is a rising interest in less conservative statistical testing procedures which enjoy increased statistical power, such as FDR control~\cite{Benjamini95}. Another critical problem is how to correct for confounders~\cite{Epstein12}. Those are predictive features which are correlated to both the target response and some of the patterns, artificially inflating the resulting $p$-values. Extending the framework in either of those directions would represent a very valuable contribution. 

\balance


\begin{small}
\bibliographystyle{abbrv}
\bibliography{bibliography}  
\end{small}


\end{document}